%% file: ieee_journal_main.tex
\documentclass[lettersize,journal]{IEEEtran}
\input{preamble}

\usepackage{amsmath,amsfonts,amsthm}
\usepackage{algorithmic}
\usepackage{array}
\usepackage{textcomp}
\usepackage{stfloats}
\usepackage{url}
\usepackage{verbatim}
\usepackage{graphicx}
\usepackage{xcolor}
\def\BibTeX{{\rm B\kern-.05em{\sc i\kern-.025em b}\kern-.08em
    T\kern-.1667em\lower.7ex\hbox{E}\kern-.125emX}}
\usepackage{balance}

\newtheorem{lemma}{Lemma}
\newcommand\defeq{\mathrel{\stackrel{\makebox[0pt]{\mbox{\normalfont\scriptsize def}}}{:=}}}

\begin{document}
\raggedbottom
\title{Revisiting Out-of-Distribution Detection in Real-time Object Detection: From Benchmark Pitfalls to a New Mitigation Paradigm}

\author{Changshun Wu$^{1,*}$, Weicheng He$^{1,*}$, Chih-Hong Cheng$^{2,3}$, Xiaowei Huang$^{4}$, and Saddek Bensalem$^{5}$ \\ 
\thanks{*Equal contribution.}
\thanks{$^{1}$ Universit\'e Grenoble Alpes, Grenoble, France.}
\thanks{$^{2}$ Carl von Ossietzky University of Oldenburg, Oldenburg, Germany.}
\thanks{$^{3}$ Chalmers University of Technology, Gothenburg, Sweden.}
\thanks{$^{4}$ University of Liverpool, Liverpool, UK.}
\thanks{$^{5}$ CSX-AI, Grenoble, France.}
\thanks{Correspondence to: changshun.wu@univ-grenoble-alpes.fr}
}

\markboth{IEEE Journal of \LaTeX\ Template}%
{How to Use the IEEEtran \LaTeX \ Templates}

\maketitle

\input{sections/0_abstract}


\input{sections/1_introduction}

\input{sections/2_related_work}

\input{sections/3_benchmark_analysis}
\input{sections/4_approach}

\input{sections/5_experiments}
\input{sections/6_in-depth_analysis}

\input{sections/7_discussion}
\input{sections/8_conclusion}




\bibliographystyle{ieeetr}
\bibliography{ref_long}

\begin{IEEEbiography}[{\includegraphics[width=1in,height=1.25in,clip,keepaspectratio]{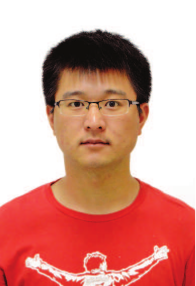}}]{Changshun Wu} received his PhD in Automation from Aix-Marseille University in 2019. He is currently a researcher fellow at Universit\'e Grenoble Alpes and the Verimag Laboratory in France. His research focuses on developing safe and interpretable AI systems by combining interpretability methods, formal verification for safety guarantees, robustness evaluation, and runtime monitoring for detecting anomalies and out-of-distribution inputs. 
\end{IEEEbiography}

\begin{IEEEbiography}[{\includegraphics[width=1in,height=1.25in,clip,keepaspectratio]{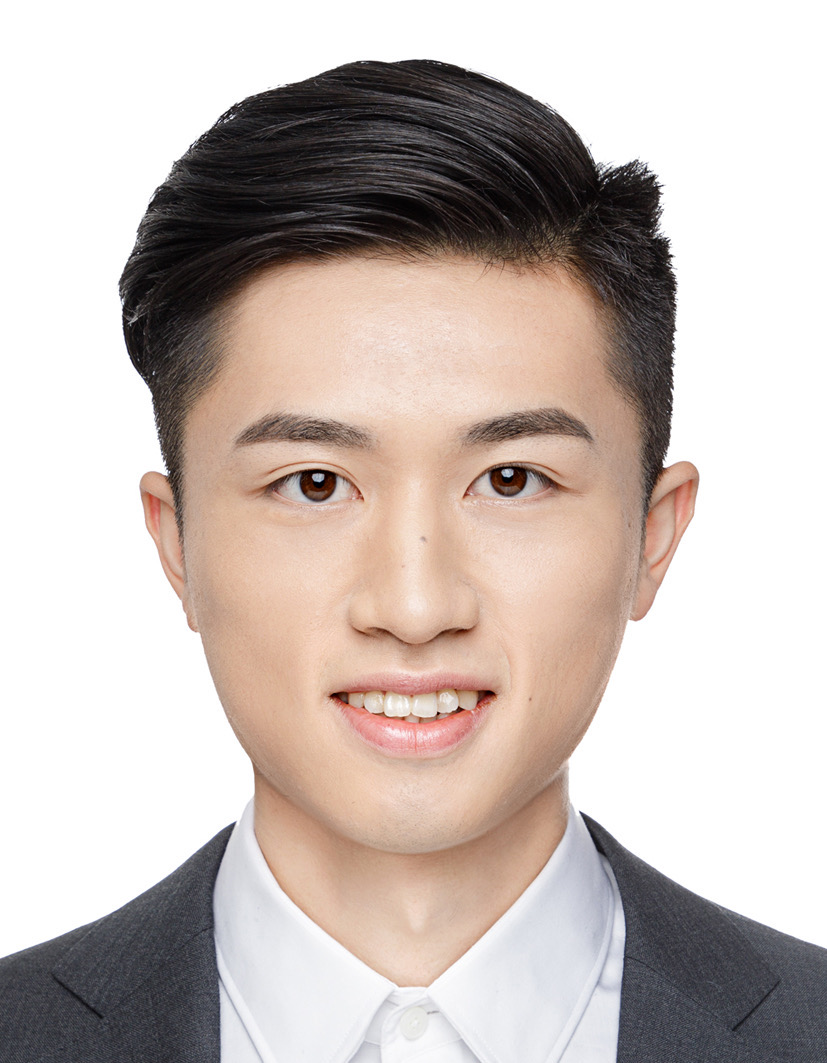}}]{Weicheng He} is currently a Ph.D. candidate in computer science at the University of Grenoble Alpes. Before this, he received his M.S. degree in 2020 from the University of Lorraine in France and his B.S. degree in 2018 from Nanjing University of Aeronautics and Astronautics in China. His research interests include out-of-distribution detection and computer vision.
\end{IEEEbiography}

\begin{IEEEbiography}[{\includegraphics[width=1in,height=1.25in,clip,keepaspectratio]{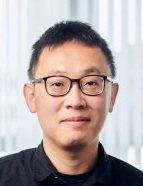}}]{Chih-Hong Cheng} received the M.Sc. degree in electrical engineering from the National Taiwan University, Taipei, Taiwan, in 2008, and the Ph.D. in informatics from the Technical University of Munich, Garching, Germany, in 2012. Currently, he is a professor at the University of Oldenburg, Germany. Before Oldenburg, he was a tenured faculty member at Chalmers University of Technology, Sweden, where he still works part-time. His research interests include software engineering, formal methods, and AI/ML for trustworthy autonomy. After his PhD, he mainly worked in government (fortiss, Fraunhofer IKS) and industrial (ABB, DENSO) research centers while holding interim professorships (TU Munich, Hildesheim) in Germany. 
\end{IEEEbiography}

\begin{IEEEbiography}
[{\includegraphics[width=1in,clip,keepaspectratio]{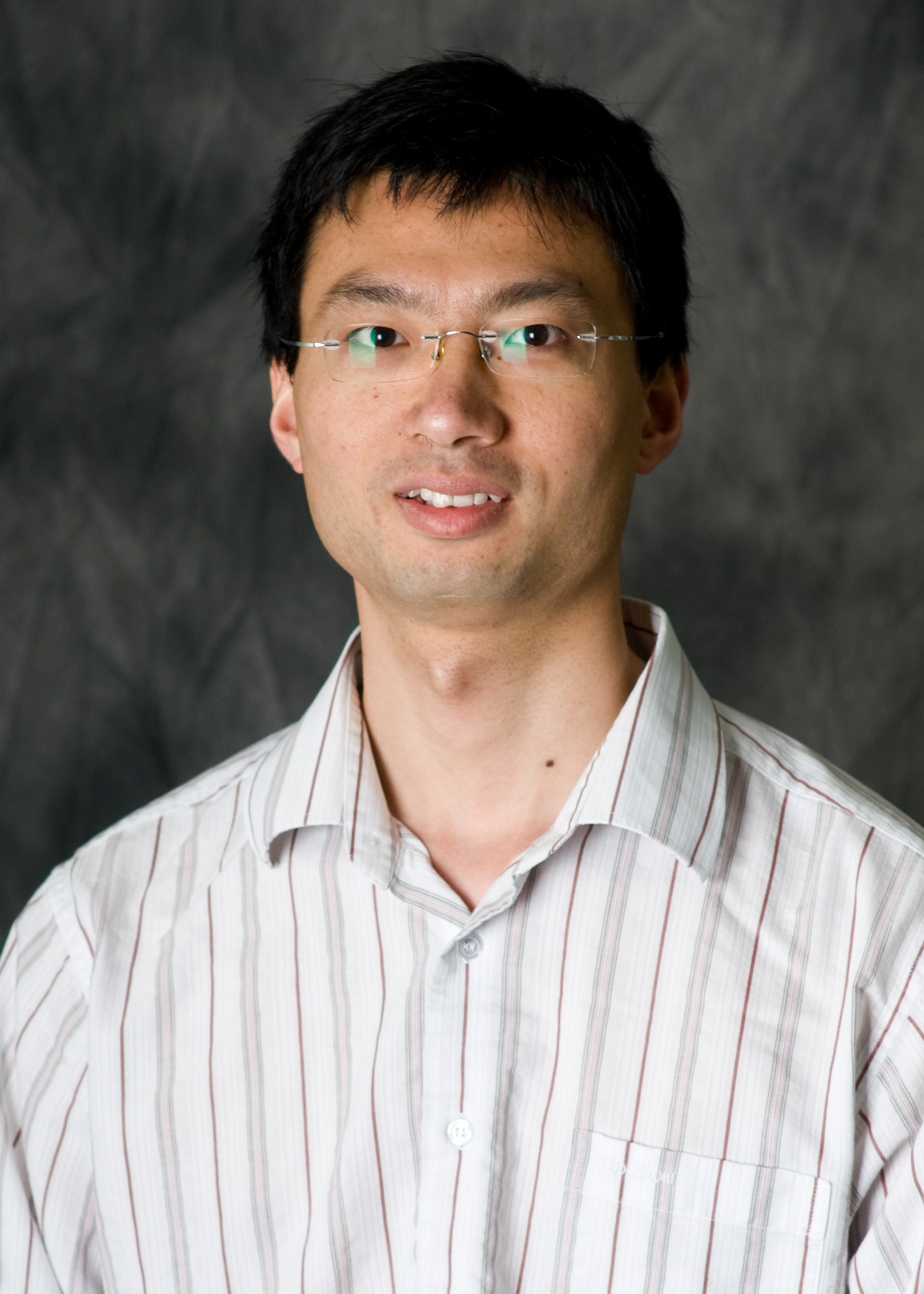}}]{Xiaowei Huang} is a Professor of Computer Science at the University of Liverpool, where he founded the Trustworthy Autonomous Cyber-Physical Systems Lab. He works on AI and machine learning algorithms with a focus on developing rigorous methods to improve the safety, security, and trustworthiness of deep learning models. His research explores verification, falsification, improvement, and explanation techniques for deep learning. 
\end{IEEEbiography}

\begin{IEEEbiography}[{\includegraphics[width=1in,height=1.25in,clip,keepaspectratio]{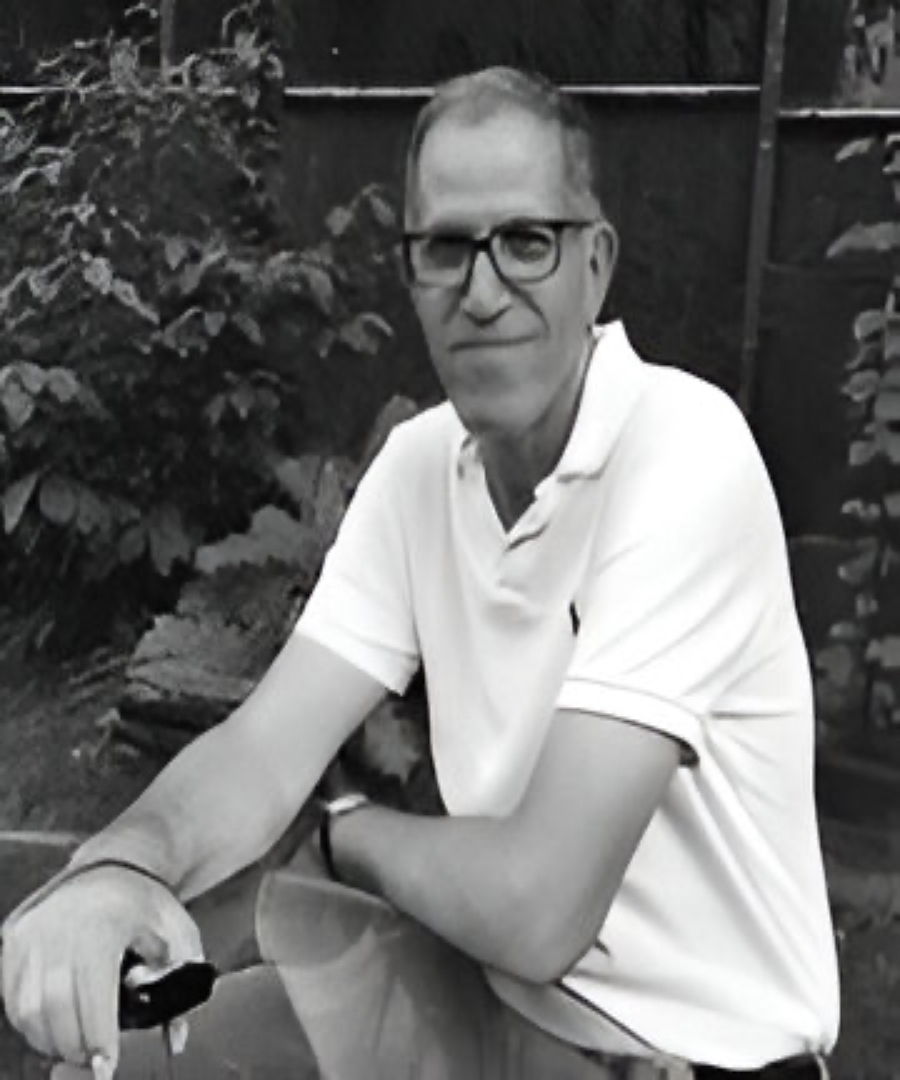}}]{Saddek Bensalem} is the CEO of CSX-AI, France. He previously was a professor at Universit\'e Grenoble Alpes, where he led research on rigorous system design and verification methods. His work centered on developing scalable verification tools and formalizing the design process to ensure correctness from the outset. His current research interests include safe AI, with an emphasis on robustness, interpretability, and out-of-distribution input detection.
\end{IEEEbiography}

\end{document}

%% file: preamble.tex
\usepackage{graphicx}
\usepackage{caption}
\usepackage{subcaption}
\usepackage{multirow}
\usepackage{url} 
\usepackage{booktabs}
\usepackage{color}
\usepackage{makecell}
\usepackage{array}
\usepackage{balance}
\usepackage{hyperref}
\usepackage[utf8]{inputenc}
\usepackage{graphicx}

\newcommand{\Comment}[1]{}

%% file: sections/0_abstract.tex
\begin{abstract}
Out-of-distribution (OoD) inputs pose a persistent challenge to deep learning models, often triggering overconfident predictions on non-target objects.
While prior work has primarily focused on refining scoring functions and adjusting test-time thresholds, such algorithmic improvements offer only incremental gains.
We argue that a rethinking of the entire development lifecycle is needed to mitigate these risks effectively.
This work addresses two overlooked dimensions of OoD detection in object detection.
First, we reveal fundamental flaws in widely used evaluation benchmarks: contrary to their design intent, up to 13\% of objects in the OoD test sets actually belong to in-distribution classes, and vice versa.
These quality issues severely distort the reported performance of existing methods and contribute to their high false positive rates.
Second, we introduce a novel training-time mitigation paradigm that operates independently of external OoD detectors.
Instead of relying solely on post-hoc scoring, we fine-tune the detector using a carefully synthesized OoD dataset that semantically resembles in-distribution objects.
This process shapes a defensive decision boundary by suppressing objectness on OoD objects, leading to a 91\% reduction in hallucination error of a YOLO model on BDD-100K.
Our methodology generalizes across detection paradigms such as YOLO, Faster R-CNN, and RT-DETR, and supports few-shot adaptation.
Together, these contributions offer a principled and effective way to reduce OoD-induced hallucination in object detectors.
Code and data are available at: \url{https://gricad-gitlab.univ-grenoble-alpes.fr/dnn-safety/m-hood}. 
\end{abstract}

%% file: sections/1_introduction.tex
\section{INTRODUCTION}\label{sec:intro}
Deep learning models, such as image classifiers and object detectors, have achieved remarkable success in accomplishing in-distribution (ID) tasks by learning meaningful patterns.
However, these models tend to overgeneralize when exposed to \emph{out-of-distribution (OoD)} inputs, i.e., samples that fall outside the training distribution and do not correspond to any ID category. Predictions on OoD inputs often result in confident yet nonsensical or irrelevant outputs.  
We refer to this failure mode as \emph{hallucination} in computer vision.
The risks associated with such OoD-induced hallucinations are twofold.
They may arise unintentionally, as in \emph{type errors} or \emph{semantic bugs} in code, where the syntax is correct but the meaning is wrong.
Alternatively, they can be exploited deliberately, allowing adversaries to induce high-confidence false detections without requiring perturbations or access to training data.

\begin{figure}[t]
    \centering
    \begin{subfigure}{0.44\textwidth}
        \centering
        \includegraphics[width=0.98\textwidth]{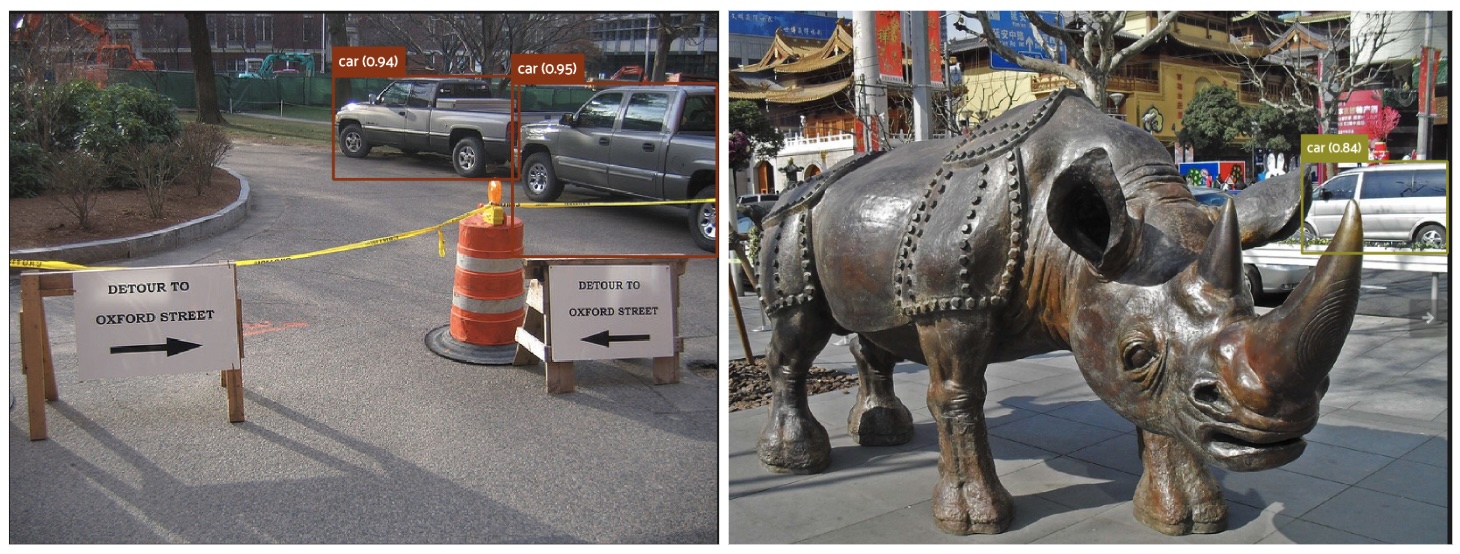}  
        \caption{Test images used in VOS~\cite{du2022towards} that should be diagnosed as OoD accidentally contain ID ``car'' objects.}
        \label{fig:id_in_ood}
    \end{subfigure}
    \hfill
    \begin{subfigure}{0.44\textwidth}
        \centering
        \includegraphics[width=0.98\textwidth]{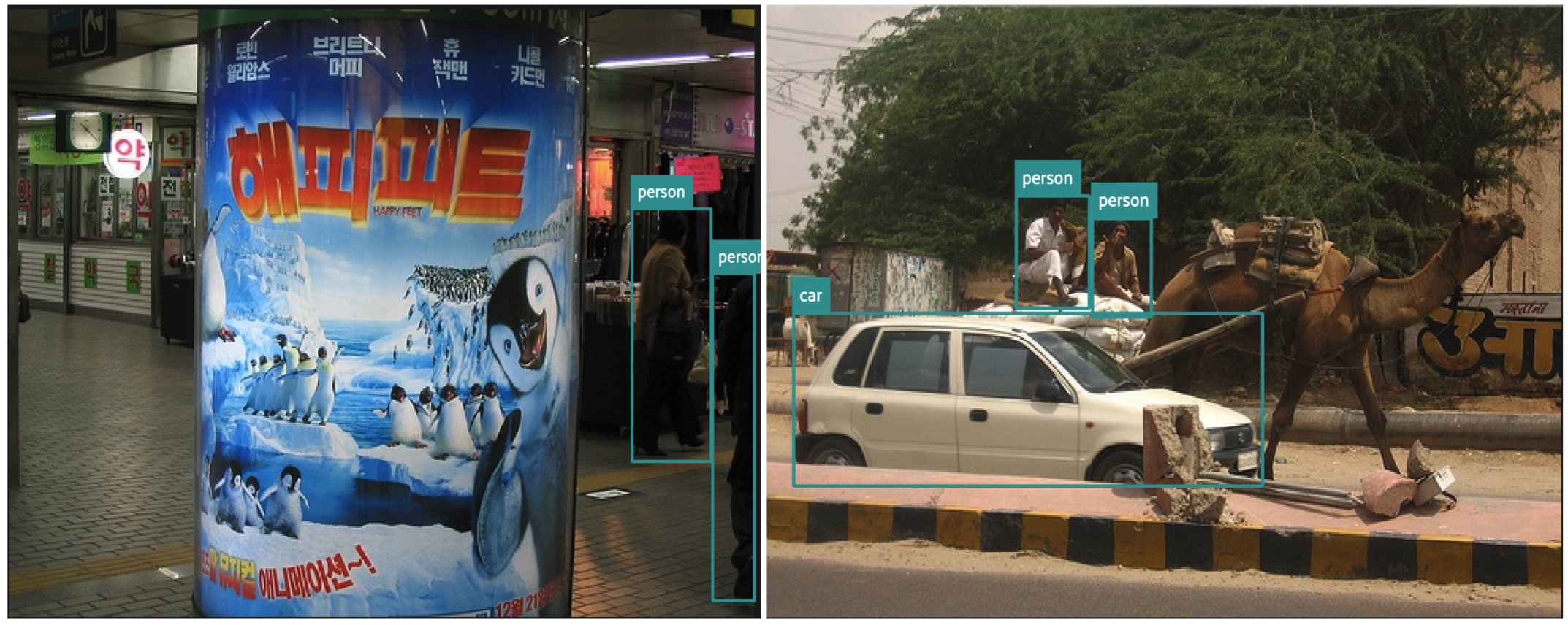}  
        \caption{ID images used in VOS~\cite{du2022towards} for shaping the decision boundary containing OoD objects such as penguins and camels.}
        \label{fig:ood_in_id}
    \end{subfigure}
    \caption{Quality issues in existing OoD evaluation benchmarks that lead to non-optimal performance}
    \label{fig:dataset_examples}
\end{figure}

\begin{figure*}[t]
    \centering
    \includegraphics[width=0.95\textwidth]{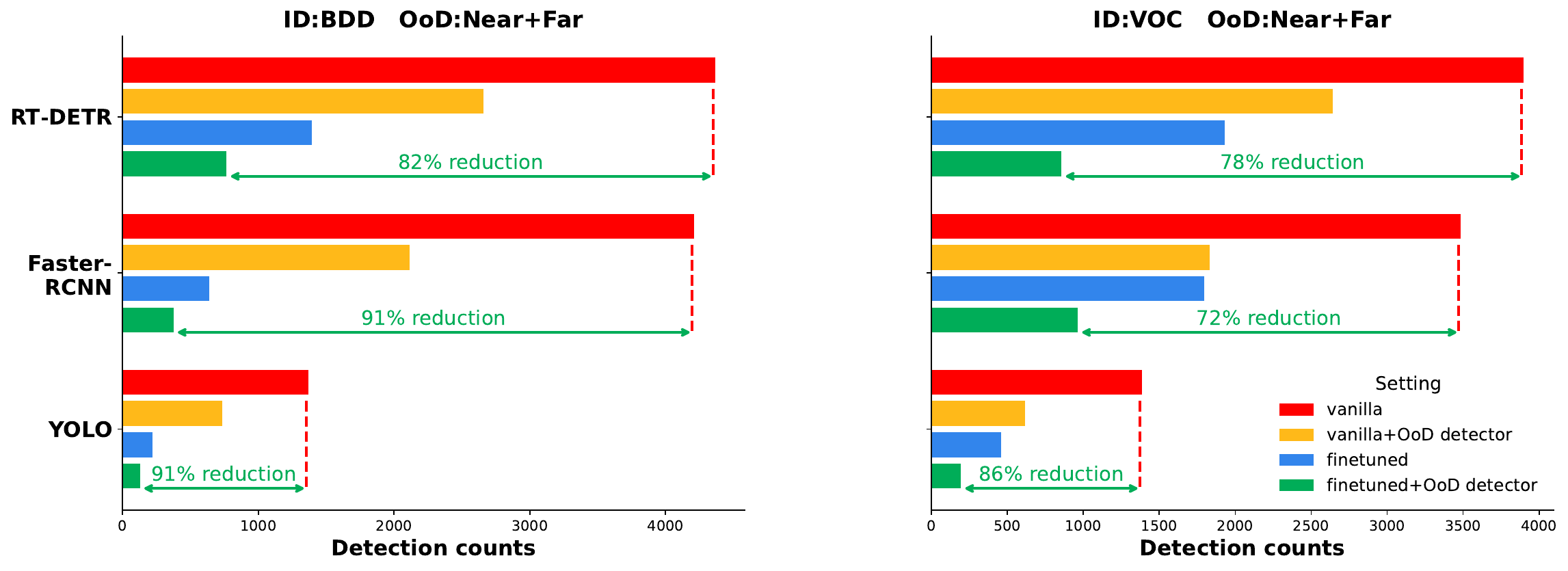}
    \caption{Reduction of hallucination errors on OoD datasets. Our approach significantly reduces hallucinated detections on OoD inputs across two ID datasets (BDD100K and PASCAL-VOC) and multiple detection architectures (YOLO, Faster R-CNN, RT-DETR). The method achieves substantial reductions in spurious detections—up to 91\%—and generalizes well across different detector families, including one-stage, two-stage, and Transformer-based models.}
    \label{fig:method_highlight}
\end{figure*}

To mitigate the risks posed by such illegal inputs, OoD detection has emerged as a promising solution, attracting extensive research efforts focused on designing effective test-time filtering algorithms.
For instance, in image classification, these methods typically assign low confidence scores to OoD samples and redirect their predictions to a dedicated ``OoD'' class.
Inspired by this paradigm, recent techniques such as VOS~\cite{du2022towards}, SAFE~\cite{wilson2023safe}, and BAM~\cite{wu2024bam} are implemented on top of Faster-RCNN~\cite{ren2015faster} as an extra component to subsequently filter the detected and classified objects by setting these objects to be of class ``OoD''.

While prior work has largely focused on refining OoD scoring functions and adjusting test-time thresholds, we argue that such algorithmic efforts for OoD detection, though valuable, offer only incremental progress in the context of object detection. 
Unlike classification tasks, which produce a single prediction per image, object detection involves structured outputs with multiple objects and spatial reasoning, making the detection and mitigation of OoD instances considerably more complex.
A recent study~\cite{he2024box} illustrates this challenge by adapting leading classification-based OoD techniques to YOLO detectors, only to find a substantial drop in performance compared to their application on Faster R-CNN~\cite{wu2024bam}.
These findings suggest that post-hoc algorithms alone are insufficient.
We contend that optimizing OoD detection performance cannot rely on a single algorithmic solution.
Instead, it calls for a re-examination of the entire development lifecycle of current mitigation paradigms in object detection.
In this work, we address two underexplored aspects: (1) fundamental implementation flaws in existing OoD detection benchmarks, and (2) the lack of training-time mitigation strategies that explicitly leverage the structural properties of object detectors, independent of external OoD filters.

First, we identify \emph{dataset quality issues and benchmark design} in evaluating OoD used in VOS~\cite{du2022towards} and all other techniques, contributing to the high false positive rates observed in previous studies.
Specifically, we uncover two critical problems and \emph{study their impact both in theory and concrete empirical evaluations}: 
\begin{itemize}
    \item Existing OoD test datasets have many ``OoD'' classified images. However, these images contain objects holding in-distribution (ID) classes, i.e., categories defined in the object detector training dataset. Consider the example in Fig.~\ref{fig:id_in_ood} where an image is considered as ``OoD'', \emph{any} prediction of a ``car'' object in that image will be counted as a hallucination error. Due to the dataset design and the evaluation pipeline, an image's ``OoD'' labeling is only on the \emph{whole image level} (i.e., not on the bounding box level). Such a labeling creates an \emph{implicit assumption} that an ``OoD'' image should not contain any ID object. However, as shown in later sections, around~$13\%$ of objects detected in ``OoD'' images are actually ID. This implies that the performance for prior results is higher than the reported numbers by removing dataset errors. 
    \item For constructing the decision boundary of filters, existing frameworks use \textit{FPR95} (false positive rate at $95\%$ true positive rate) as a guiding principle. 
    The decision boundary threshold is adjusted so that $95\%$ of the distribution samples are correctly classified. The ID images used to calibrate decisions also contain potentially diagnosed OoD objects counted as errors (see Fig.~\ref{fig:ood_in_id} for examples), so the intended decision boundary can be influenced, leading to worse performance. 
\end{itemize}

Second, to reduce hallucinations (overconfident predictions) caused by OoD samples, we propose a new mitigation paradigm that forges a defensive decision boundary against OoD inputs through fine-tuning.
Unlike classification models, object detectors are inherently trained to differentiate foreground objects from background objects, often using the \emph{objectness score}.
As OoD filtering is only triggered when the object detector acknowledges the existence of an object, reducing the false detection can be done as a \emph{joint approach}: In addition to equipping a filter similar to that of VOS or BAM, additionally adjust the decision boundary of the object detector to reduce opportunities to trigger false detections. 
Specifically, we introduce a fine-tuning strategy that makes use of a set of OoD samples proximal to ID data - termed ``proximal OoD'' - which can be inferred from the ID data distribution without relying on specific test-time OoD datasets. 
From the perspective of learned ``concepts'', these samples share high-level semantic features with ID data while differing in low-level attributes, effectively bridging the gap between ID and truly novel OoD instances.
During fine-tuning of the models on these proximal OoD samples, the model is trained to treat proximal OoD samples as background, thereby shaping an implicit defensive decision boundary against general OoD instances.
Our experimental results, as shown in Fig.~\ref{fig:method_highlight}, demonstrate that the proposed method effectively mitigates hallucinations  across different experimental settings when encountering both near-OoD and far-OoD inputs, demonstrating its efficacy.

In summary, our main contributions are as follows:
\begin{itemize}
    \item We identify critical dataset quality issues in existing OoD benchmarks and develop an automated tool for constructing customized evaluation datasets tailored to an object detector’s training categories.
    \item We propose a novel fine-tuning approach that leverages objectness properties to construct a more robust defensive decision boundary against OoD samples.
    \item We show through large-scale experiments that our approach generalizes effectively across diverse detection paradigms, including single-stage CNN-based models (e.g., YOLO~\cite{wang2025yolov10}), two-stage models (e.g., Faster R-CNN~\cite{ren2015faster}), transformer-based detectors (e.g., RT-DETR~\cite{zhao2024detrs}), and few-shot fine-tuning settings.
    \item We provide additional insights into the effectiveness of our method through explainability and confidence shift analysis.
\end{itemize}

The remainder of this paper is organized as follows: Section~\ref{sec:rw} reviews related work in OoD detection for image classification and object detection. Section~\ref{sec:ood.problem.benchmark} details our analysis of dataset quality issues. Section~\ref{sec:finetuning} presents our proposed fine-tuning approach. Section~\ref{sec:exp} reports the experimental results, and Section~\ref{sec:analysis} analyzes the effectiveness of our approach. Section~\ref{sec:discussion}  discusses the open challenges in mitigating OoD inputs for object detection. Section~\ref{sec:conclusion} concludes the paper with discussions on future directions.

%% file: sections/2_related_work.tex
\section{RELATED WORK}\label{sec:rw}
\begin{figure*}[t]
    \centering
    \includegraphics[width=0.98\linewidth]{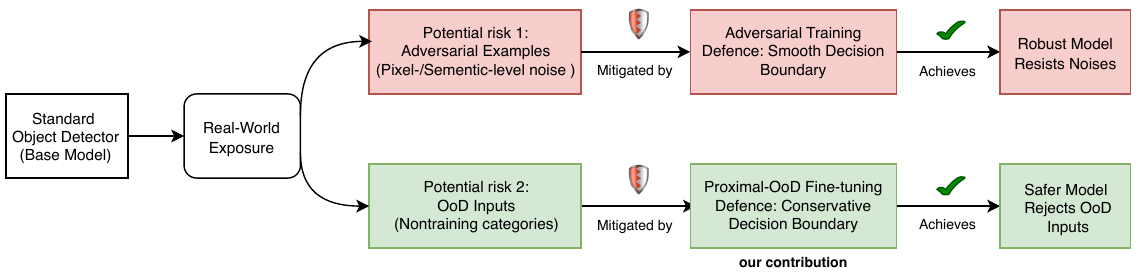}
    \caption{Positioning of our training-time OoD safety method (green) within the broader defense landscape. Adversarial defenses (red) improve robustness against malicious perturbations through adversarial training and inference-time defenses. In parallel, our method (green) proactively shapes decision boundaries to handle risky OoD inputs. This previously overlooked training-time safety mechanism complements post hoc filtering and aligns with a broader AI safety paradigm: combining robustness and reliability across the training–inference pipeline.}
    \label{fig:methodPosition}
\end{figure*}

Overconfidence or hallucination is a symptom that can arise in both ID and OoD scenarios, where our work focuses on the OoD case. OoD detection has gained significant attention in computer vision. Within image classification tasks, techniques leveraging confidence scores~\cite{liang2018enhancing,liu2020energy,xu2024scaling}, feature-space distances~\cite{lee2018simple,cheng2019runtime,sun2022out,wu2023customizable} or generative modeling~\cite{ren2019likelihood,kamkari2024geometric} can detect OoD anomalies. We refer readers to a recent survey paper~\cite{yang2024generalized} for a detailed summary of techniques. Due to the maturity of OoD detection research in classification, the community has developed a standard benchmark OpenOoD~\cite{yang2022openood}, enabling researchers to compare against different techniques. For OoD detection in object detection, the problem is less explored despite its importance, and as of the time of writing (March 2025), OpenOoD has not yet supported object detection. Methods operated on Faster-RCNN, including VOS~\cite{du2022towards}, SAFE~\cite{wilson2023safe} and BAM~\cite{wu2024bam} enable reducing hallucination in OoD samples via refining feature space representations and uncertainty estimation. 
The recent study~\cite{he2024box} translated multiple OoD detection methods within classification to the YOLO-family object detection settings, including maximum logits (MLS)~\cite{hendrycks2022scaling}, maximum softmax probability (MSP)~\cite{hendrycks2017baseline}, energy-based methods (EBO)~\cite{liu2020energy}, Mahalanobis distance (MDS)~\cite{lee2018simple}, k-nearest neighbors (KNN)~\cite{sun2022out} and the recent SCALE method~\cite{xu2024scaling}. While the performance of reducing hallucination against OoD samples has been considerably superior to those applied on Faster-RCNN, it is still far from satisfactory. 
Our result in highlighting problems within the existing object detection OoD benchmark dataset implies that the performance of prior results should be better than the reported numbers. It also stresses the importance of having a rigorous benchmark to ensure an unbiased performance evaluation.  

Furthermore, our fine-tuning strategy—based on injecting proximal OoD examples during training—marks a shift in methodology within the realm of OoD detection for object detectors. Prior efforts have largely focused on designing inference-time filters to reject OoD objects.
In contrast, our approach leverages the unique structure of object detectors—particularly the objectness score—to proactively shape safer decision boundaries during training.
This constitutes a training-time safety mechanism that complements traditional post hoc filtering.
Such a perspective aligns with a broader trend in machine learning safety: just as adversarial robustness~\cite{tramer2019adversarial} research has evolved to include both adversarial training and inference-time defenses, we argue that OoD safety should be treated as a joint endeavor across the training–inference pipeline.
As illustrated in Fig.~\ref{fig:methodPosition}, adversarial training improves resilience against malicious, fine-grained perturbations, while our method improves reliability under semantically incompatible inputs.
Together, these complementary defenses help characterize a more holistic safety framework for open-world object detection.

This work builds upon our earlier conference version~\cite{he2025mitigating}, which focused on YOLO-based object detectors. We substantially extend that study by (i) developing an automated OoD benchmarking tool, (ii) evaluating multiple detection paradigms including Faster R-CNN and RT-DETR, (iii) exploring few-shot fine-tuning strategies, and (iv) providing deeper analyses of interpretability and effectiveness.

%% file: sections/3_benchmark_analysis.tex
\section{OOD PROBLEM AND BENCHMARK ANALYSIS}\label{sec:ood.problem.benchmark}


\textit{Existing Benchmark.} Existing benchmarks for evaluating OoD detection in object detection, including those by existing results~\cite{wilson2023safe,wu2024bam,he2024box} continue to utilize the datasets introduced in the work of VOS~\cite{du2022towards}.
These benchmarks claim that the OoD test datasets do not contain any object that overlaps with one of the ID classes.
However, we have identified critical issues within these datasets that contribute to inflated FPR95 values, ultimately undermining the accuracy of the evaluation.
In the following, we formulate the observed phenomenon and analyze these issues in detail.

\subsection{Issues in Benchmark Datasets - Theoretical Analysis}\label{subsec:mislabeling}  

Let $f$ be an object detector that takes an image $I$ and produces a set $f(I) \defeq \{ (b_i, y_i) \}$ of bounding box $(b_i)$ and the associated label $(y_i)$. The label $y_i$ is within the list of ID object classes $\mathcal{O}_{\text{in}}$. In the example of applying~$f$ under the KITTI dataset~\cite{geiger2012we}, 
$\mathcal{O}_{\text{in}}$ includes classes such as \texttt{car}, \texttt{van}, or \texttt{truck}. For OoD detection, it is achieved by furthering applying a \emph{filtering function} $g(b_{i}, I)$ to each predicted object $b_{i}$ in image $I$, where $g$ outputs a score that reflects the confidence of the object not within the ID classes. 
A threshold $\tau$ is then applied to the score such that if $g(b_{i}, I) < \tau$, the object is classified as ID, and the label $y_i$ is kept the same. Otherwise, it is classified as OoD and $y_i$ is updated to a special class $\texttt{ood}$, i.e., $y_i \gets \texttt{ood}$ where $\texttt{ood} \not \in \mathcal{O}_{\text{in}}$.

\subsubsection{How OoD-only test sets are evaluated}  

Given an image~$I$, let $G_{\mathcal{O}_{\text{in}}}(I)  \defeq \{ (\hat{b}_i, \hat{y}_i) \}$ be the associated ground truth object label associated with~$\mathcal{O}_{\text{in}}$. 
An \emph{OoD-only test set} $\mathcal{D}_{\text{test}}^{\text{OoD}}$ contains only images~$I$ where $G_{\mathcal{O}_{\text{in}}}(I) = \emptyset$, i.e., there exists no object $\hat{b}_i$ whose true class~$\hat{y}_i$ is within~$\mathcal{O}_{\text{in}}$. 
To evaluate the performance of the detector~$f$ and the OoD filter~$g$ against $\mathcal{D}_{\text{test}}^{\text{OoD}}$, one iteratively takes $I \in \mathcal{D}_{\text{test}}^{\text{OoD}}$, generate predictions and apply filtering. As $G_{\mathcal{O}_{\text{in}}}(I) = \emptyset$, the number of hallucinations made in~$f$ combined with~$g$ can be summarized using Eq.~\eqref{eq:error.sum.fp}. It is the number of generated bounding boxes whose labels are not changed to $\texttt{ood}$. 

\begin{multline}\label{eq:error.sum.fp}
err^{+}_{f,g}(I) \defeq |\{(b_i, y_i) \; | \; (b_i, y_i) \in f(I), g(b_i, I) <  \tau\}| \\
\text{provided that} \;\; I \in \mathcal{D}_{\text{test}}^{\text{OoD}} 
\end{multline}

\subsubsection{Type~1 data error: ID Objects in OoD Test Sets}  

We can now describe the Type~1 data error and its associated consequences. A Type~1 error occurs when an image in the OoD-only test set contains at least one ID object (i.e., belonging to \(\mathcal{O}_{\text{in}}\)). An example of this issue is shown in Fig.~\ref{fig:id_in_ood}.
Precisely, an \emph{OoD-only test set} $\mathcal{D}_{\text{test}}^{\text{OoD}}$ manifests \textbf{Type~1 data errors} if and only if the following holds:

\begin{equation}\label{eq:type.1.error}
\exists I \in \mathcal{D}_{\text{test}}^{\text{OoD}} \;\; \text{such that} \;\; G_{\mathcal{O}_{\text{in}}}(I) \neq \emptyset
\end{equation}

Consequently, if $f$ is very successful in detecting objects specified in $\mathcal{O}_{\text{in}}$, the impact of data error will be more likely to be reflected in the evaluation to be counted as the false positive made by~$f$ and~$g$. 

\begin{lemma} Given $I \in \mathcal{D}_{\text{test}}^{\text{OoD}}$ where $G_{\mathcal{O}_{\text{in}}}(I) \neq \emptyset$, and assume that $f$ detects a bounding box with success probability $\alpha$. The expected value of false positives (hallucinated objects) made by $f$ and~$g$, caused by the Type~1 data error, equals $\alpha |G_{\mathcal{O}_{\text{in}}}(I)|$.
\end{lemma}

\proof The lemma holds as every ID-object in $G_{\mathcal{O}_{\text{in}}}(I)$ within an OoD-only image can be detected by $f$ with a probability of $\alpha$. 

\subsubsection{How ID-only data set is used for deciding FPR95 decision boundary for the OoD filter}  

To set the threshold~$\tau$ for deciding whether an object shall be considered as OoD or ID, the criterion of FPR95 (false positive rate at $95\%$ true positive rate) is used. This means that $\tau$ is adjusted on the ID-only dataset $\mathcal{D}_{\text{cali}}^{\text{ID}}$ where the OoD filter achieves a~$95\%$ success rate in ``not to consider an object as OoD''. Counting the errors made in an image within~$\mathcal{D}_{\text{cali}}^{\text{ID}}$ is based on Eq.~\eqref{eq:error.sum.fn}. 
We use an example to assist in understanding Eq.~\eqref{eq:error.sum.fn}: If there is a car in the image where the object detector successfully detects it (i.e., $(b_i, y_i) \in f(I)$), but the OoD filter negatively considers it an OoD object (i.e., $g(b_i, I) \geq  \tau$), it should be counted as an error of~$g$. 

\begin{equation}\label{eq:error.sum.fn}
    \begin{aligned}
err^{-}_{f,g}(I) \defeq |\{(b_i, y_i) \; | \; (b_i, y_i) \in f(I), g(b_i, I) \geq  \tau\}|  \\\;\;\;\;\;
 \text{provided that} \;\; I \in \mathcal{D}_{\text{cali}}^{\text{ID}} 
    \end{aligned}
\end{equation}

Based on the FPR95 principle, one then carefully adjusts the threshold~$\tau$ such that within $\mathcal{D}_{\text{cali}}^{\text{ID}}$, total error rate $err_{\mathcal{D}_{\text{cali}}^{\text{ID}}}$ remains less than $5\%$, where $err_{\mathcal{D}_{\text{cali}}^{\text{ID}}}$ is defined using Eq.~\eqref{eq:error.total}.

\begin{equation}\label{eq:error.total}
err_{\mathcal{D}_{\text{cali}}^{\text{ID}}} \defeq  \frac{\sum_{I \in \mathcal{D}_{\text{cali}}^{\text{ID}}} err^{-}_{f,g}(I)}{\sum_{I \in \mathcal{D}_{\text{cali}}^{\text{ID}}}  |\{(b_i, y_i) \; | \; (b_i, y_i) \in f(I)\}|} 
\end{equation}

\subsubsection{Type~2 data issue: Unlabelled OoD Objects in ID-only data sets}\label{subsec:type.2}  

Unfortunately, the above experiment design is based on an assumption where \emph{there exist no OoD objects} that an object detector~$f$ can be wrongly triggered. 
Precisely, an \emph{ID-only dataset} $\mathcal{D}_{\text{cali}}^{\text{ID}}$ manifests \textbf{Type~2 data issues} if and only if the following holds, where $\hat{y}_i$ is the associated ground truth:

\begin{equation}\label{eq:type.2.error}
\exists I \in \mathcal{D}_{\text{cali}}^{\text{ID}} \;\; \text{such that} \;\; \exists (b_i, y_i) \in f(I): \hat{y}_i = \texttt{ood} 
\end{equation}

Consider the images illustrated in Fig.~\ref{fig:ood_in_id}, where they belong to $\mathcal{D}_{\text{cali}}^{\text{ID}}$. Consider if an object detector~$f$ wrongly detects the camel as an object within class~$\mathcal{O}_{\text{in}}$, and the filter $g$ successfully filters it. Then, the condition in Eq.~\eqref{eq:error.sum.fn} holds, so it will be counted as an error made by the OoD filter~$g$. 

In the following, we explain the impact on the decision boundary~$\tau$ to be integrated into the OoD filter~$g$ by considering one construction method $L_2$-norm-based OoD filter. Techniques such as BAM or Mahalanobis distance share the same characteristic, so the proof strategy can be easily adapted to other OoD filters.  

Given an object detector $f$ with input $I \in \mathcal{D}_{\text{cali}}^{\text{ID}}$, let $f^{-l}(b_i, I)$
be the feature vector at layer~$l$ that generates bounding box~$b_i$. Let $c$ be the arithmetic mean of all feature vectors for all boxes in $\mathcal{D}_{\text{cali}}^{\text{ID}}$, i.e.,
\begin{equation}
  c \defeq\frac{\sum_{I \in \mathcal{D}_{\text{cali}}^{\text{ID}}} \sum_{(b_i, y_i) \in f(I)} f^{-l}(b_i, I)}{\sum_{I \in \mathcal{D}_{\text{cali}}^{\text{ID}}} \sum_{(b_i, y_i) \in f(I)} 1} 
\end{equation}

\begin{figure}[ht]
    \centering
        \includegraphics[width=0.5\columnwidth]{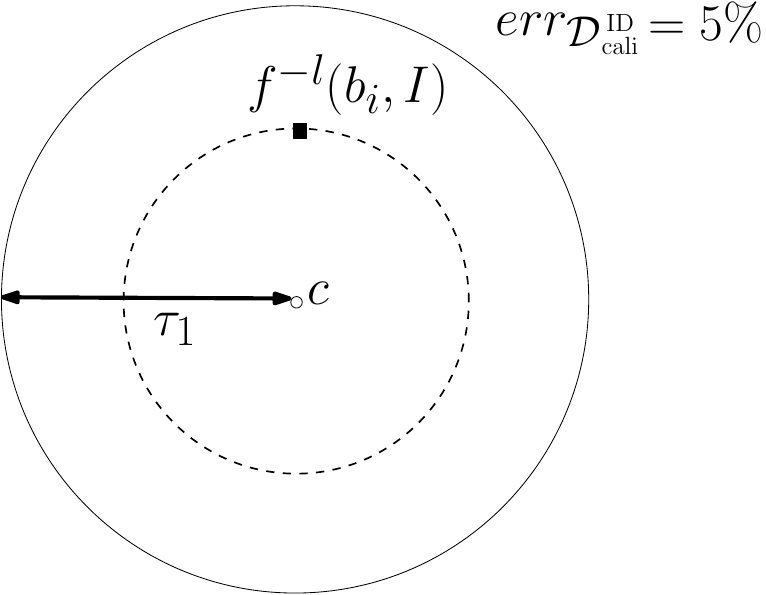}  
        \caption{Understanding the impact of decision boundary when unlabeled OoD object occurs in an ID-only dataset}
        \label{fig:proof}
\end{figure}

Consider an OoD filter $g(b_i, I) \defeq ||f^{-l}(b_i, I), c \,||_2$ that computes the $L_2$ norm between the feature vector and the center~$c$. Therefore, $\tau$ is defined based on the distance from the feature vector to the center (mean). Consider using FPR95 to decide~$\tau$, where the construction is illustrated using Fig.~\ref{fig:proof}. One sorts the feature vectors by distances (to center) from large to small. Subsequently, start by setting~$\tau$ to be the largest distance~$\tau_{max}$, and greedily introduce a feature vector with the next smaller distance~$\tau'$ and update $\tau$ to~$\tau'$. When adjusting $\tau$, ensure that $err_{\mathcal{D}_{\text{cali}}^{\text{ID}}}$ remains to be below~$5\%$.  Let~$\tau_1$ in Fig.~\ref{fig:proof} be the radius where the currently computed error rate is exactly~$5\%$. We discuss the impact of including~$f^{-l}(b_i, I)$ in the design of~$\tau$, where $b_i$ contains an OoD object (triggering Type~2 data issue).
\begin{itemize}

    \item Based on the standard evaluation pipeline, then shrinking~$\tau$ from $\tau_1$ to $g(b_i, I)$ makes the condition $g(b_i, I) \geq \tau$ in Eq.~\eqref{eq:error.sum.fn} from \texttt{false} to \texttt{true}, implying that the resulting error rate $err_{\mathcal{D}_{\text{cali}}^{\text{ID}}}$ (with small radius) will be increased, thereby being larger than~$5\%$. 

\item If one changes the evaluation pipeline where bounding box~$b_i$ can be evaluated against the ground-truth label ``$\texttt{ood}$'', the prediction over $b_i$ will be counted as correct. Consequently, shrinking~$\tau$ from $\tau_1$ to $g(b_i, I)$ still keeps $err_{\mathcal{D}_{\text{cali}}^{\text{ID}}}$ to be~$\leq 5\%$. 
    
\end{itemize}

Therefore, having an image with a bounding box containing an OoD object in the ID dataset (without labeling these objects as ``\texttt{ood}'' and enabling an evaluation pipeline) increases the length of~$\tau$. As a larger~$\tau$ implies weaker capabilities in filtering OoD objects (recall in Fig.~\ref{fig:proof}, where everything inside the circle is counted as ID by filter~$g$), the capability of distance-based OoD filters, when there exist unlabeled OoD objects within the ID dataset, can be compensated. 

\vspace{-3mm}

The following summarizes our theoretical analysis;  in the subsequent sections, we provide a quantitative analysis of the concrete benchmark suites supporting our findings.  

\begin{itemize}

\item Fixing Type~1 data issues requires carefully removing ID objects within an OoD-only dataset. 
    \item Fixing Type~2 data issues requires labeling all potentially identified OoD objects in bounding boxes (to let an OoD detector's response be counted as correct) within an ID-only dataset.\footnote{Alternatively, design a pipeline that every object outside the labelled ones is automatically counted as OoD.}
\end{itemize}

\subsection{Issues in Benchmark Datasets - Quantitative Analysis}
We quantitatively analyze Type~1 error in the widely used OoD test dataset, OpenImages~\cite{OpenImages}, designed for evaluating OoD detection methods for two ID tasks: PASCAL-VOC~\cite{everingham2010pascal} and BDD-100K~\cite{yu2020bdd100k}. 
For brevity, we refer to this subset as OpenImages throughout this paper.
We focus on instances where ID objects are erroneously included in the benchmark OoD-only test set, potentially distorting evaluation metrics.
We leverage a high-accuracy pre-trained object detection model trained on OpenImages V7~\cite{OpenImages}, which recognizes over 500 categories.
By running inference on the entire OpenImages dataset, we systematically identify images containing ID objects that should not have been classified as OoD.
Our analysis reveals that among the 1,852 images in the dataset, 104 images (5.62\%) contain at least one ID object despite being labeled as OoD.
Across these images, we detect 220 ID objects, accounting for 12.75\% and 13.77\% of all detected objects by the models for PASCAL-VOC and BDD-100K tasks (see Table~\ref{tab:mislabel_impact}).
To measure the impact of these mislabeling errors on OoD detection performance, we evaluate FPR95 with and without the erroneously labeled images.
Our results in Table~\ref{tab:mislabel_impact} show that Type~1 error inflates FPR95 by~$9.85$ percentage points and~$5.58$ percentage points in PASCAL-VOC and BDD-100K tasks, respectively, demonstrating a significant distortion in evaluation metrics.
These findings underscore the need for more rigorous dataset curation to ensure the reliability of OoD detection benchmarks.

\begin{table}[ht]
    \centering
    \caption{Quantitative analysis of Type~1 error: Estimation of the prevalence of ID object in OoD test set and its impact on OoD detection evaluation.}
    \label{tab:mislabel_impact}
    \renewcommand{\arraystretch}{1.2}
    \resizebox{0.48\textwidth}{!}{
    \begin{tabular}{lccc}
    \hline 
        \textbf{ID Dataset} & \textbf{OoD Dataset} & \textbf{Prevalence} & \textbf{FPR95 Inflation} \\
        \hline
        PASCAL-VOC & OpenImages & 12.75\% & +9.85\% \\
        BDD-100K & OpenImages & 13.77\% & +5.58\% \\ \hline
    \end{tabular}
    }
\end{table}


Given the large number of predictions—nearly 200,000 in BDD-100K and nearly 20,000 in PASCAL-VOC—it is impractical to assign each prediction an accurate ID/OoD label.
Therefore, for the Type~2 issue, we approximately quantify its impact on evaluation through a pre-trained model by estimating the number of OoD objects present in the ID test sets of the PASCAL-VOC and BDD-100K tasks.
As shown in Table \ref{tab:type2}, 16.52\% of images in the BDD-100K dataset and 8.69\% in the PASCAL-VOC dataset contain at least one OoD object, which is non-negligible. 
%
%

\begin{table}[t]
    \centering
    \caption{Quantitative analysis of Type~2 issues: Estimation of ID and OoD object counts in the BDD-100K and PASCAL-VOC datasets using a pre-trained model.}
    \label{tab:type2}
    \renewcommand{\arraystretch}{1.2}
    \resizebox{0.4\textwidth}{!}{ 
    \begin{tabular}{lccc}
        \hline
        \textbf{ID Dataset} & \textbf{Type} & \textbf{Objects} & \textbf{Samples} \\
        \hline
        \multirow{2}{*}{BDD-100K} & ID  & 185,945  & 10,000   \\
        & OoD & 8,129  & 1,652 (16.52\%)  \\
        \hline
        \multirow{2}{*}{PASCAL-VOC} & ID  & 12,648  & 4,952   \\
        & OoD & 746  & 430 (8.69\%)  \\
        \hline
    \end{tabular}
    }
\end{table}


\subsection{ID Outlier-induced Ambiguity}
\begin{figure}[ht]
    \centering
    \includegraphics[width=0.96\columnwidth]{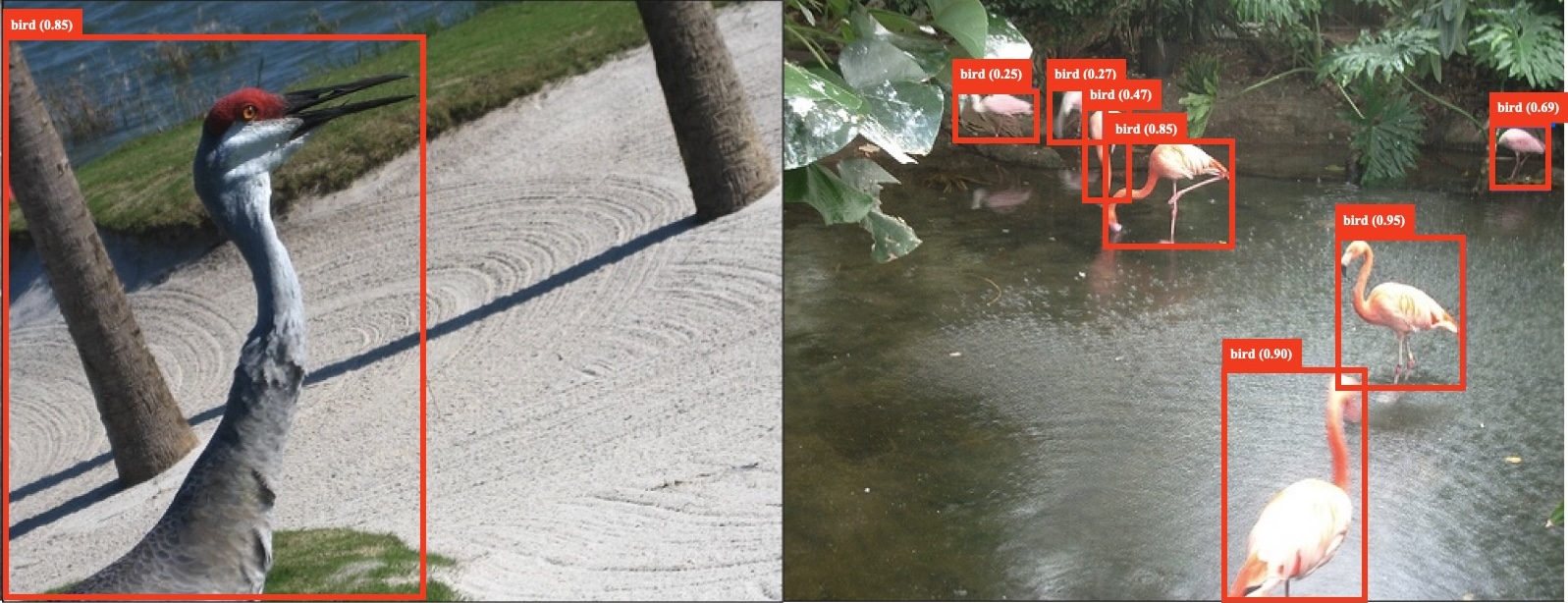} %
    \caption{Visualization of ID outliers in the ``Bird" category in PASCAL-VOC: atypical bird samples with rare shapes or colors that deviate from the main category distribution.}
    \label{fig:viz_outlier}
\end{figure}

\begin{figure}[ht]
    \centering
    \includegraphics[width=0.415\textwidth]{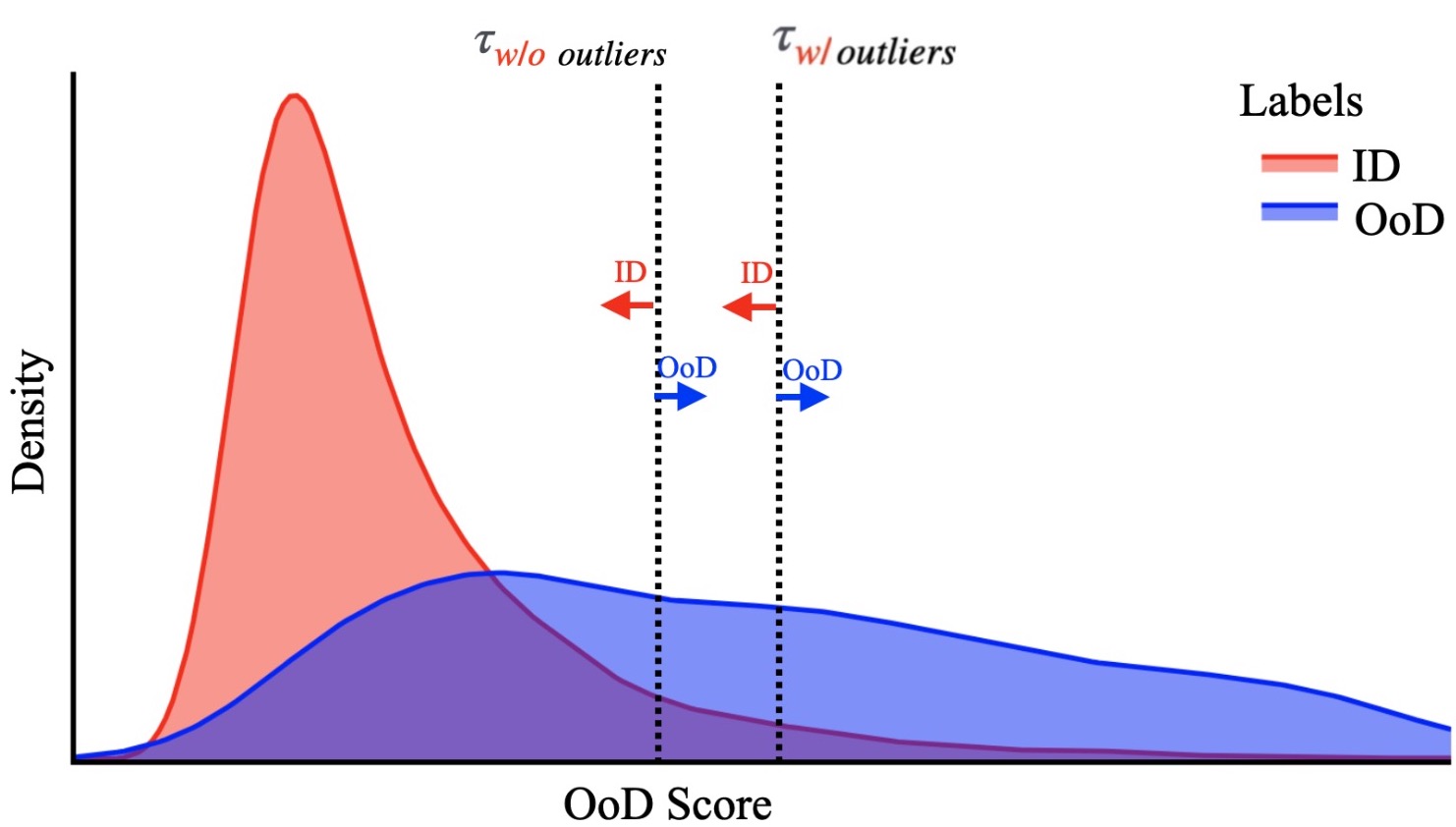} %
    \caption{Kernel Density Estimation (KDE) plot of the OoD scores for a KNN-based OoD filter, on a benchmark where the ID task is PASCAL-VOC and the OoD test dataset is OpenImages. A higher score indicates a higher likelihood of being an OoD object. The thresholds $\tau_{\text{w/ outliers}}$ and $\tau_{\text{w/o outliers}}$ are chosen to ensure a 95\% recall for ID detections. The presence of outliers shifts the threshold, leading to more OoD samples being misclassified as ID samples, thereby reducing the effectiveness of OoD detection.}
    \label{fig:kde_plot}
\end{figure}

Apart from Type~1 and Type~2 data issues, we identify another potential reason why OoD detection methods struggle in YOLO models: the existence of ``ID outlier objects'' (i.e., visually atypical ID objects in some particular ID categories) within the ID test dataset.
For instance, Fig.~\ref{fig:viz_outlier} shows ``ID outlier objects'' in the ``Bird'' category of PASCAL-VOC, including visually atypical birds.
Our experiments reveal that these ID outlier objects often receive exceptionally high OoD scores—sometimes even higher than most actual OoD samples.
As shown in Fig.~\ref{fig:kde_plot}, the presence of ID outliers increases the threshold $\tau$ required to retain 95\% of ID detections (analogous to the result in Sec.~\ref{subsec:type.2}). 
As a result, under this inflated OoD score threshold, many OoD samples are incorrectly classified as ID objects, meaning the OoD detector fails to filter them out effectively, leading to a high FPR95 value.
%
%

Considering the presence of these outliers, we advocate evaluating OoD filter methods on both the original dataset and a version with outliers removed.
For instance, in Table~\ref{tab:performance_comparison_openimages}, we present a comparison showing that OoD filter methods perform significantly better without these outliers.
To filter outliers, we employ a distance-based identification technique~\cite{sun2022out} that uses box plots to detect potential outliers within each category.
%

\begin{table}[t]
\centering
\caption{Comparison of OoD filters, with and without outliers, on a benchmark where the ID task is PASCAL-VOC, and the OoD test dataset is OpenImages. In the absence of outliers, these filters perform better, as reflected in lower FPR95 scores.}
\renewcommand{\arraystretch}{1.2}
\resizebox{\columnwidth}{!}{
\begin{tabular}{cccccccc}
\hline
\textbf{Condition} & \textbf{MSP} & \textbf{EBO} & \textbf{MLS} & \textbf{MDS} & \textbf{SCALE} & \textbf{BAM} & \textbf{KNN} \\ \hline
w/ outliers  & 0.7164 & 0.9307 & 0.9220 & 0.8457 & 0.6803 & 0.5306 & 0.5376 \\
w/o outliers & \textbf{0.6797} & \textbf{0.9301} & \textbf{0.9208} & \textbf{0.8410} & \textbf{0.6296} & \textbf{0.4479} & \textbf{0.4467} \\
\hline
\end{tabular}
}
\label{tab:performance_comparison_openimages}
\end{table}

\subsection{Automated Data Curation Pipeline}\label{sec:pipeline}

Both our benchmarking and fine-tuning strategies depend on high-quality, task-specific image data. However, manually collecting such data is labor-intensive—particularly when adapting to diverse sources or meeting the specific demands of evaluation and fine-tuning.
To address this challenge, we introduce a universal automated data curation pipeline that streamlines the construction of customized datasets across different experimental stages.
This pipeline integrates the FiftyOne library~\cite{moore2020fiftyone} with the YOLOE ~\cite{wang2025yoloerealtimeseeing} open-vocabulary object detection model to systematically gather and curate relevant images while ensuring the exclusion of ID content.

\begin{figure}[t]
    \centering
    \includegraphics[width=0.65\columnwidth]{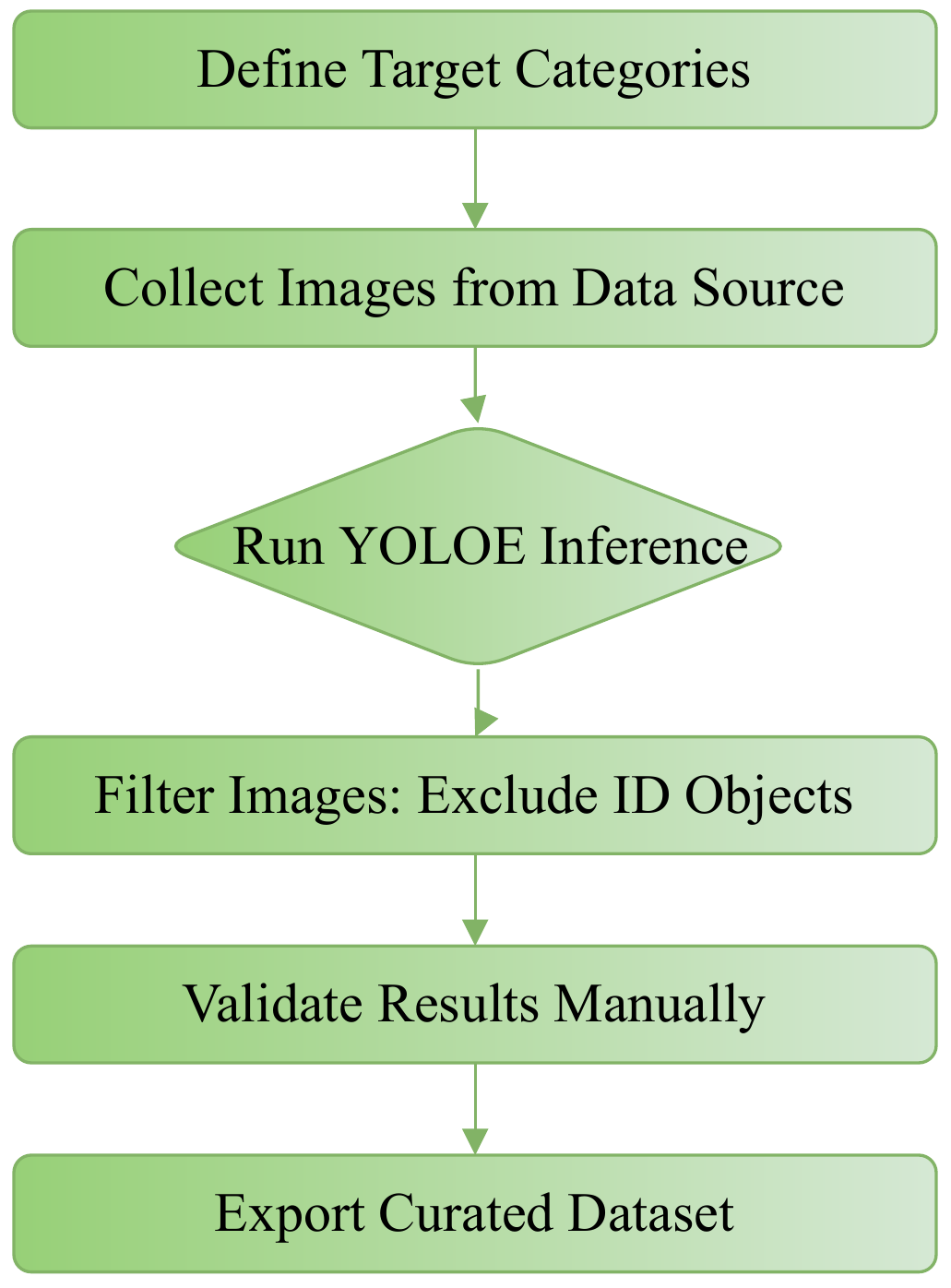}
    \caption{Overview of our universal pipeline for automated data curation.}
    \label{fig:pipeline}
\end{figure}

As illustrated in Fig.~\ref{fig:pipeline}, the pipeline workflow begins with users defining task-specific categories tailored to their requirements, whether for fine-tuning, testing, or few-shot learning scenarios.
Leveraging the FiftyOne Dataset Zoo API, it then retrieves candidate images for these categories from selected data sources, with configurable per-category image counts to accommodate various experimental needs. 
An automated YOLOE-based filter is then applied to remove ID contamination.
For each candidate image, YOLOE runs with the ID class names as prompts, and the image is discarded if any ID class is detected above the fixed thresholds: confidence $\geq$ 0.25 and NMS IoU $\geq$ 0.45.
To maintain dataset purity, the pipeline retains only images that are free from predicted or annotated ID objects.
For additional quality assurance, an optional human-in-the-loop verification step is available through FiftyOne's interactive visualization and annotation tools.
The final curated dataset is exported in standard format, ensuring seamless integration with downstream training and evaluation pipelines.
This automated and modular approach streamlines the construction of high-quality object detection datasets, maintaining consistency and reproducibility regardless of the underlying data source or specific task requirements.

\subsection{New Benchmarks}\label{sec:new_benchmarks}
To address the limitations identified in existing OoD benchmarks, we leverage our automated data curation pipeline~\footnote{\url{https://gricad-gitlab.univ-grenoble-alpes.fr/dnn-safety/m-hood/-/tree/main/src/methods/ood_benchmark_curation}}to construct two new evaluation datasets: Near-OoD and Far-OoD~\footnote{\url{https://huggingface.co/datasets/HugoHE/m-hood-dataset}}, specifically designed to enable fine-grained analysis of hallucination suppression.
Near-OoD comprises images containing objects that are visually and semantically similar to the training categories, presenting a particularly challenging scenario for object detectors.
In contrast, Far-OoD contains images with objects that are distinctly different from the training set, as well as backgrounds without recognizable objects.
Each benchmark contains 1,000 images, carefully sampled from over 500 diverse categories in OpenImages V7.
These new benchmark datasets allow for a more reliable and challenging evaluation for OoD detection for object detection models.

%% file: sections/4_approach.tex
\section{FINE-TUNING FOR MITIGATING HALLUCINATION}\label{sec:finetuning}
\begin{figure*}
    \centering
    \includegraphics[width=0.92\linewidth]{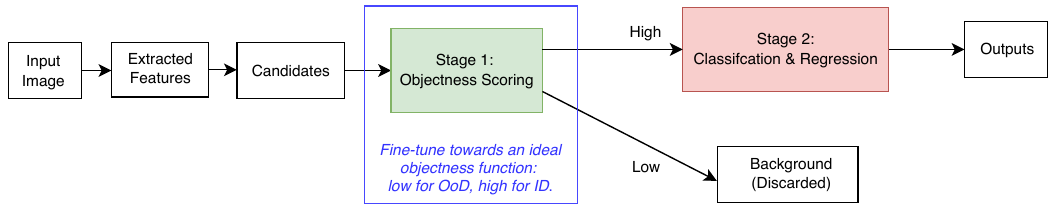}
    \caption{Conceptual two-stage decision-making logic in object detection. Regardless of architectural design (e.g., YOLO, Faster R-CNN, RT-DETR), modern detectors follow a common logical process: first, candidate regions are scored by objectness to filter out background or ambiguous inputs; second, high-scoring regions are classified and localized. Our method fine-tunes the objectness scoring mechanism using proximal OoD data, encouraging low objectness scores for OoD inputs and enabling early suppression prior to class prediction.}
    \label{fig:logicflow}
\end{figure*}

This section introduces an alternative strategy to mitigate YOLO's hallucination (overly confident predictions) on OoD samples without explicitly performing OoD detection.
Our approach leverages proximal OoD samples, which are semantically similar to the ID categories but do not belong to them.
This fine-tuning method, illustrated in Fig.~\ref{fig:logicflow}, similar to adversarial training, adjusts the model's decision boundaries to make it more conservative when facing OoD samples.
This reduces the likelihood of hallucinated predictions, whereas adversarial training focuses on enhancing the model's robustness against adversarial examples.
Rather than training a model from scratch, we adopt a fine-tuning strategy that is computationally efficient and effective in reshaping the decision boundary.
By guiding the model to differentiate between proximal OoD data and ID samples, we hypothesize that it can better generalize and distinguish broader OoD instances.
We first describe the construction of the proximal OoD dataset and then introduce our fine-tuning methodology.

\subsection{Proximal OoD Dataset Construction}\label{sec:proximal}
A well-curated proximal OoD dataset is essential for fine-tuning the model effectively to mitigate hallucinations caused by OoD samples.
We first establish selection principles for identifying suitable samples to construct such a dataset.
\begin{itemize}
    \item \textit{Non-overlapping with training categories}: Selected samples must not belong to any training categories to prevent unintentional inclusion of ID samples.
    \item \textit{Conceptual or visual similarity}: Chosen samples should be related to or visually resemble training categories, allowing the model to learn meaningful variations.
    \item \textit{Proxy category reuse}: Similar training categories may share common proxy categories when applicable. For instance, ``cat'' and ``dog'' can have overlapping proxy categories due to their similar visual characteristics.
\end{itemize}
After selecting suitable proxy categories, we use our automated data curation pipeline (see Section~\ref{sec:pipeline}) to extract approximately 2,000 images from the Caltech 256 dataset as proximal OoD samples for fine-tuning.
Caltech 256 provides a broad and diverse set of object-centric images, making it particularly suitable for capturing the subtle semantic proximity to in-distribution classes.
Importantly, sourcing data from Caltech 256 ensures no overlap or leakage with the evaluation benchmarks, thereby preserving the integrity of our experiments.

\begin{figure}[ht] 
    \centering
    \includegraphics[scale=0.18]{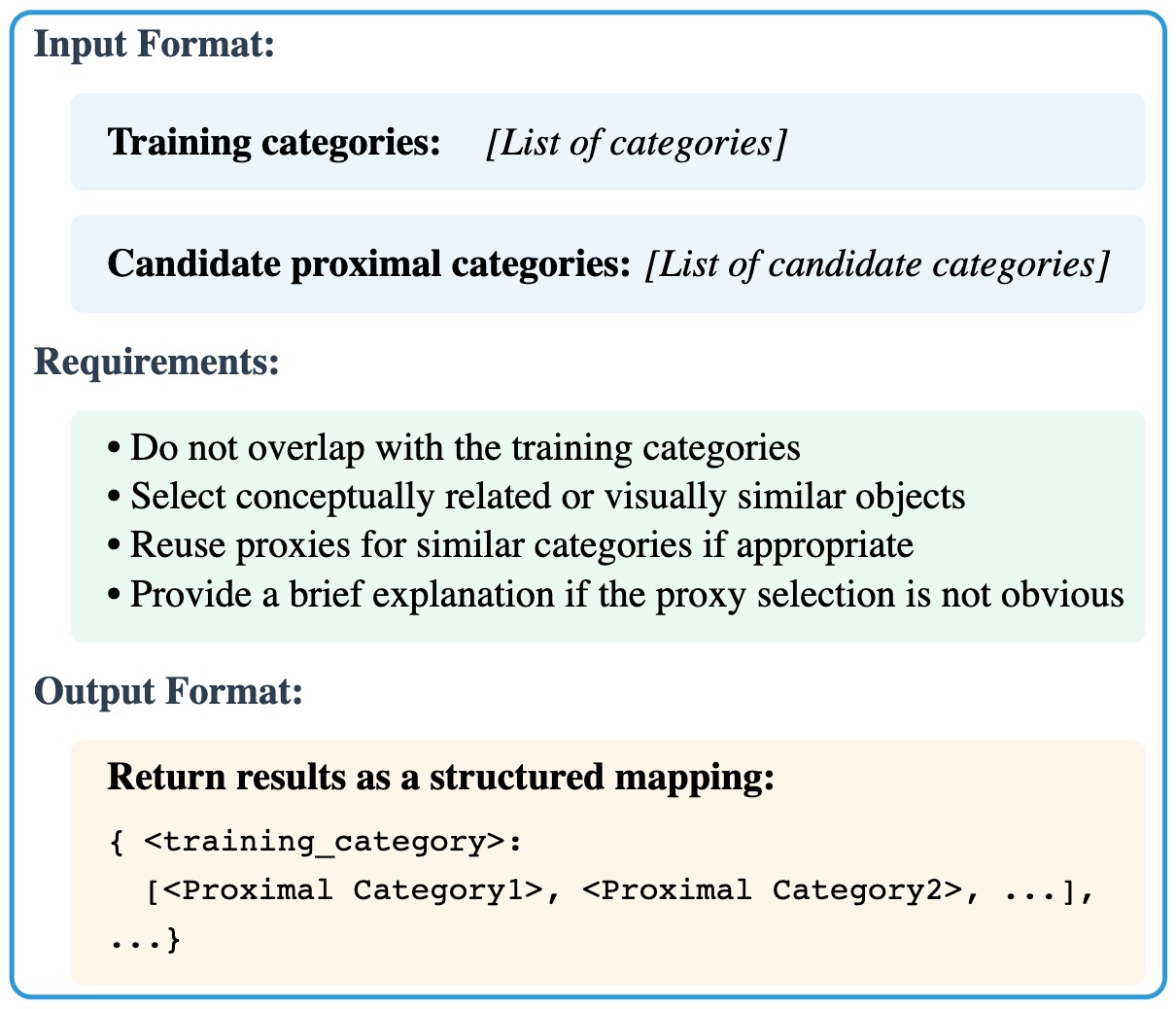} 
    \caption{Prompt template for generating candidate proximal OoD categories.}
    \label{fig:prompt}
\end{figure}

\subsection{Fine-Tuning Strategy}
To mitigate hallucinated predictions while preserving detection performance, we fine-tune the model using a dataset comprising both the original ID training set $\mathcal{D}_{train}$ and the proximal OoD sample set $\mathcal{D}_{prox}$ constructed in the previous subsection.
The only modification during training is treating proximal OoD samples as background, preventing the model from assigning high confidence scores.
The classification loss for ID samples remains as in YOLOv10~\cite{wang2025yolov10}, while for proximal OoD samples, all detected objects are treated as negatives (background).
The total classification loss is defined as:
\begin{align}
    \mathcal{L}_{cls}^{ft} = \mathcal{L}_{cls}^{\mathcal{D}_{train}} + \lambda \mathcal{L}_{cls}^{\mathcal{D}_{prox}}
\end{align}
where $\lambda$ controls the influence of proximal OoD data, this adjustment reduces hallucinations while maintaining detection accuracy.

From a theoretical perspective, our method can be interpreted through the lens of inductive bias~\cite{cao2022random,goyal2022inductive} and decision geometry.
Modern object detectors implicitly maintain a latent notion of objectness, which can be abstracted as a binary latent variable $z(x) \in \{0,1\}$ indicating whether an input region is treated as foreground or background.
Standard training learns a decision function $g_\theta(x)$ that partitions the input space based on ID data, leaving OoD regions underconstrained and prone to overconfident predictions. By introducing proximal OoD samples and enforcing $z(x)=0$ for them, our fine-tuning strategy imposes additional geometric constraints in these previously unconstrained regions, effectively reshaping the foreground–background boundary and providing a principled explanation for the observed suppression of hallucinations.

%% file: sections/5_experiments.tex
\section{EXPERIMENTS}\label{sec:exp}
This section details the experimental setup, evaluates the fine-tuned model's performance, and presents empirical results demonstrating the effectiveness of our fine-tuning method in reducing hallucinations caused by OoD samples.

\subsection{Experimental Setup}

We evaluate our method on two well-trained YOLOv10 models from~\cite{wang2025yolov10}, trained on PASCAL-VOC and BDD-100K, respectively.
We use the calibrated Near-OoD and Far-OoD datasets introduced in the previous subsection to assess hallucinations.
We use a confidence score threshold of 0.25 for all models, following the default setting in the YOLO series models~\cite{jocher2022ultralytics}.
The effectiveness of fine-tuning is measured by its ability to reduce hallucinations, which we quantify using OoD counts—the number of predictions on OoD data.
Ideally, this value should be zero.
However, ensuring that the model's primary detection performance is not compromised beyond mitigating hallucinations is crucial. 
To this end, we track mean average precision (mAP) to evaluate the model's accuracy on ID data.

\subsection{Fine-tuning}
We fine-tune the YOLOv10 models for 20 epochs on a combined dataset of their training data and proximal OoD samples collected in the previous section, following the default training configuration of YOLOv10 in~\cite{wang2025yolov10}. 
Specifically, we use the Stochastic Gradient Descent (SGD) optimizer with a momentum of 0.937 and a weight decay of $5 \times 10^{-4}$. The initial learning rate is set to 0.01 and decays linearly to 0.0001 throughout the fine-tuning process. 
Training is performed on a single NVIDIA V100 GPU with a batch size of 64. 

\begin{figure}[t]
    \centering
    \includegraphics[width=1.0\columnwidth]{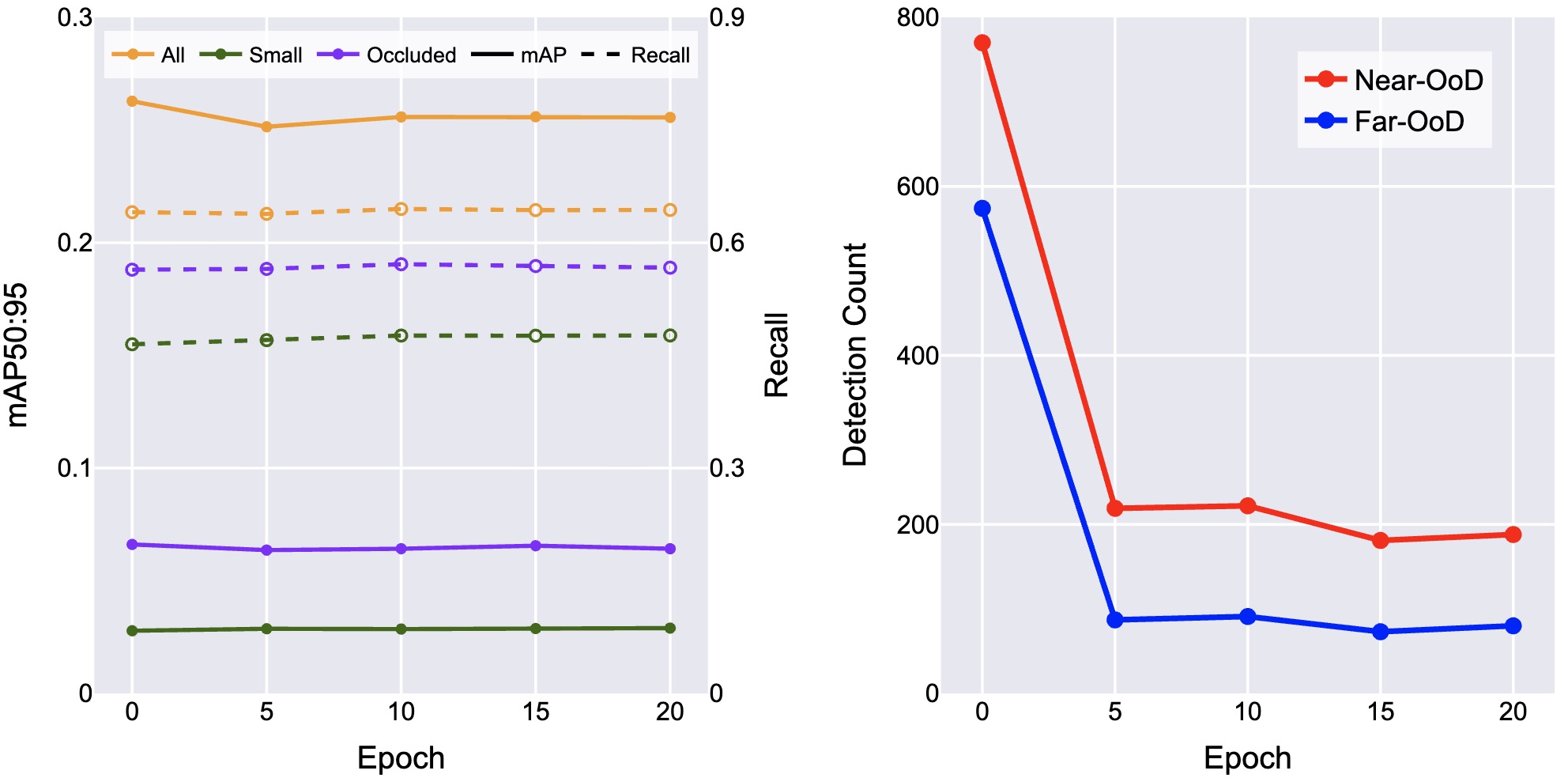}
    \caption{Fine-tuning trends of YOLOv10 model pre-trained on BDD-100K: mAP and recall evolution (left), and prediction counts on OoD datasets (right).}
    \label{fig:fine_tuning_trends}
\end{figure}

During fine-tuning, we observe the mAP on the ID validation set to analyze model performance change. 
At the same time, we evaluate the model hallucinations by measuring the number of predictions on OoD data.
Fig.~\ref{fig:fine_tuning_trends} illustrate mAP changes over epochs and the trend in hallucination counts on OoD datasets during fine-tuning for the ID task of BDD-100K. 
The results show that mAP initially declines before gradually improving and largely recovering by the end of the 20 epochs, with only a minimal decrease compared to the original baseline. 
This is likely due to the inclusion of proximal OoD data, which presents the model with more challenging backgrounds (similar to ID categories).
As it adapts, ID performance temporarily declines.
Over time, the model adapts by refining its decision boundaries, regaining ID performance while learning to handle the more challenging OoD data.
We further evaluated the model on \textit{hard ID} subsets, specifically \textit{small} objects (area $<$ 32×32 pixels) and heavily \textit{occluded} objects (Intersection-over-Area $>$ 0.5 with another box in the same image).
Fine-tuning does not lead to a noticeable degradation in either mAP or recall on these subsets, with both metrics remaining largely stable compared to the vanilla model.
This stability is likely due to the relatively low baseline performance on such challenging cases.

Meanwhile, hallucinations on both OoD datasets drop sharply in the first epoch and stabilize at a lower level after several epochs (around $20\%$ of the original level), suggesting that the model becomes more conservative early in fine-tuning, reducing overconfident predictions.
We selected a version for further analysis, reducing OoD detections by up to $87\%$ with only about $1$ percentage point decrease in ID performance. This version strikes a balance between ID task accuracy and hallucination suppression.

\subsection{Effectiveness Evaluation}
In this subsection, we evaluate the effectiveness of our fine-tuning approach in mitigating hallucinations in YOLO models when exposed to OoD samples.
Our analysis consists of two key aspects: (1) the impact of fine-tuning alone and (2) the additional benefits of integrating an OoD detector with the fine-tuned model.

\begin{figure*}[t]
    \centering
    \begin{subfigure}{\textwidth}
        \centering
        \includegraphics[width=0.9\textwidth]{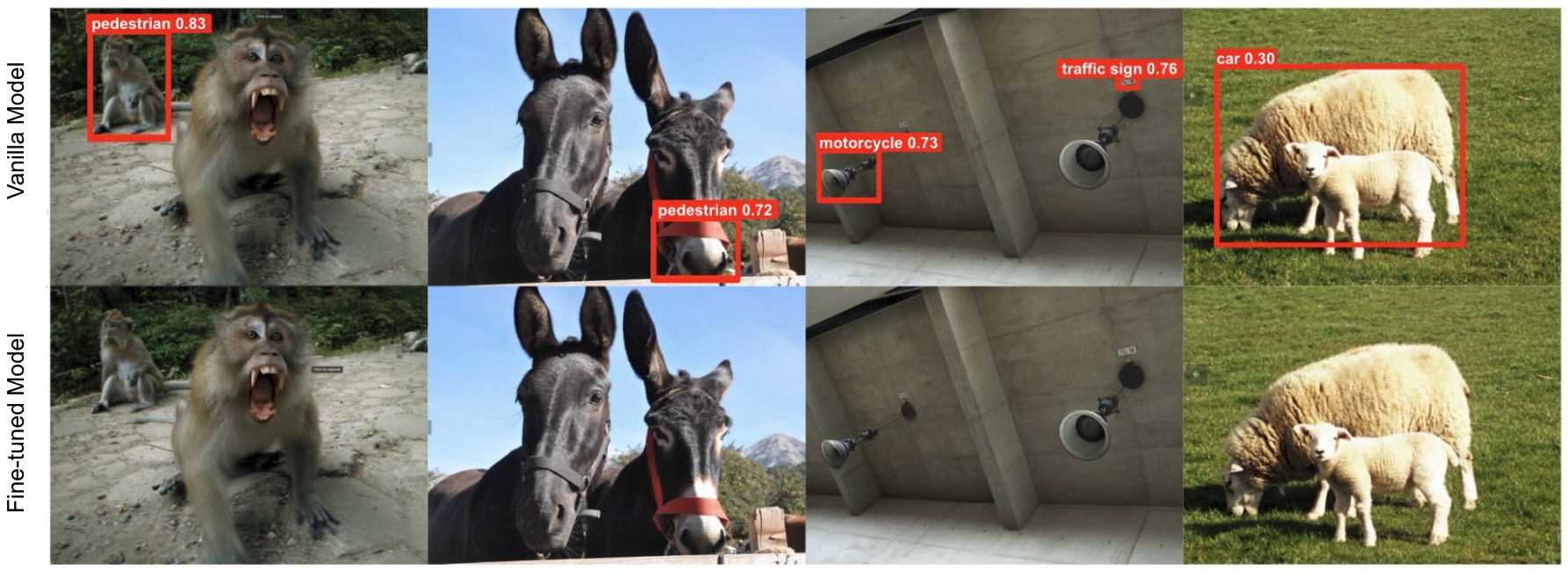}
        \subcaption{Near-OoD images}
        \label{fig:qualitative_near}
    \end{subfigure}
    
    \begin{subfigure}{\textwidth}
        \centering
        \includegraphics[width=0.9\textwidth]{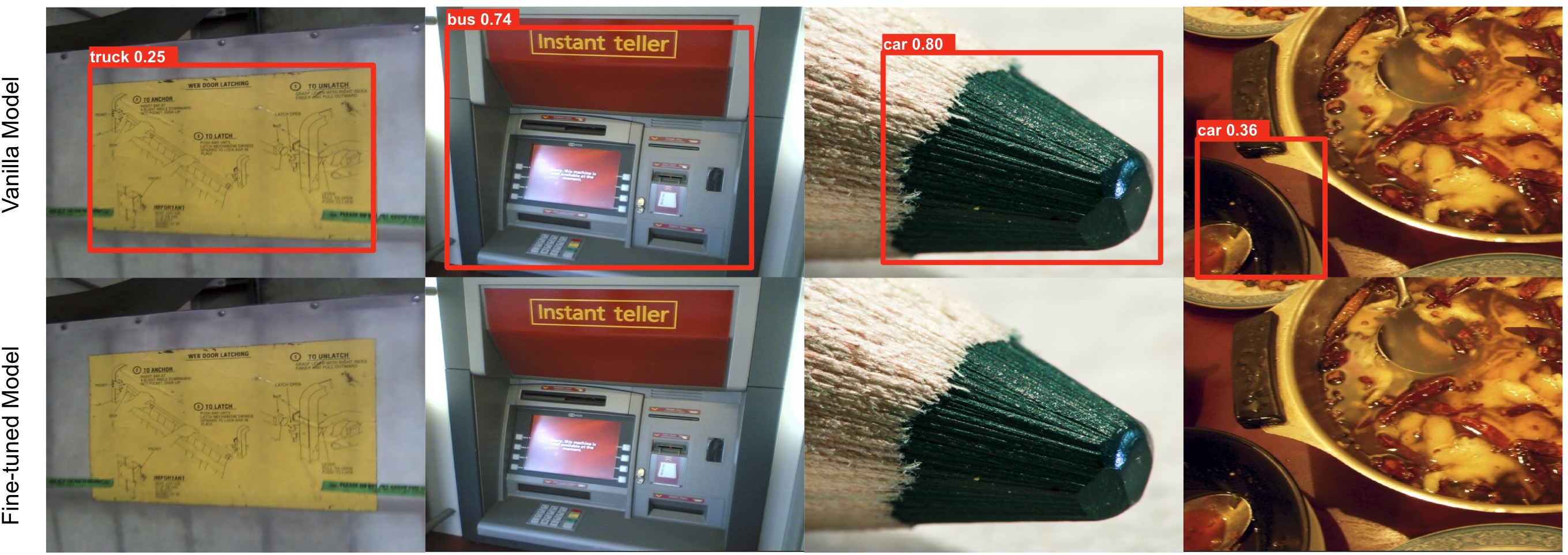}
        \subcaption{Far-OoD images}
        \label{fig:qualitative_far}
    \end{subfigure}
    \caption{Qualitative comparison of predictions on OoD images using the original and fine-tuned models. (Top) The original model produces hallucinated predictions. (Bottom) The fine-tuned model mitigates hallucinations, enhancing safety in decision-making.}
    \label{fig:qualitative_analysis}
\end{figure*}

\textbf{Impact of Fine-Tuning} To quantify the reduction in hallucinations, we compare the number of OoD detections before and after fine-tuning across different datasets, as shown in Table~\ref{tab:ood_counts}.
Fine-tuning leads to a substantial decrease in misdetections across both ID tasks.
For example, in the BDD-100K dataset, the number of detected Near-OoD samples drops from 701 to 133, and a similar reduction is observed in other settings.
This trend highlights the effectiveness of fine-tuning in suppressing spurious detections on OoD data.
A qualitative comparison is provided in Fig.~\ref{fig:qualitative_analysis}, where the fine-tuned model demonstrates fewer hallucinated predictions (highlighted in red bounding boxes) than the original model. 
This visual evidence further supports the ability of our fine-tuning approach to mitigate hallucinated predictions.

\begin{table}[ht]
\caption{OoD counts comparison of different models.}
    \label{tab:ood_counts}
    \centering
    \renewcommand{\arraystretch}{1.2}
    \resizebox{0.48\textwidth}{!}{ 
\begin{tabular}{cccccc}
    \hline
    \textbf{\makecell{ID \\ Dataset}} & \textbf{\makecell{OoD \\ Datasets}} & \textbf{Original} & \textbf{\makecell{Original+ \\KNN}} & \textbf{Fine-tuned} & \textbf{\makecell{Fine-tuned+ \\KNN}} \\
    \hline
    \multirow{2}{*}{PASCAL-VOC}
        & Near-OoD & 946 & 443 & 326 & \textbf{134} \\
        & Far-OoD & 440 & 174 & 131 & \textbf{60} \\
    \midrule
    \multirow{2}{*}{BDD-100K} 
        & Near-OoD & 701 & 363 & 133 & \textbf{80} \\
        & Far-OoD & 666 & 371 & 85 & \textbf{47} \\
    \hline
\end{tabular}
}
\end{table}

\textbf{Integrating an OoD Detector} To further mitigate hallucinations, we integrate an OoD detector into the fine-tuned model. 
Our evaluation consists of two steps: first, we assess the performance of various OoD detection methods, and second, we analyze the combined effect of fine-tuning and the best-performing detector.
Table~\ref{tab:OoDeffectiveness} presents the results of our OoD detector evaluation, where BAM and KNN consistently outperform other baselines. 
Notably, KNN achieves slightly better performance than BAM.
Given its superior performance, we use KNN combined with the original and fine-tuned models to examine their respective contributions to reducing hallucinations.
Table~\ref{tab:ood_counts} shows that fine-tuning alone significantly improves model reliability, reducing hallucination counts across all datasets. 
Notably, in the BDD-100K task, the fine-tuned model without an OoD detector significantly reduced hallucinations, with the number of Near-OoD predictions dropping from 363 (Original + KNN) to 133 and Far-OoD predictions decreasing from 371 to 85.
When KNN is combined with the fine-tuned model, hallucination counts are further reduced by up to $44.7\%$ (relative reduction), demonstrating the additional benefit of leveraging an OoD detector. 
Specifically, as shown in Table~\ref{tab:ood_counts}, on the BDD-100K benchmark, this combination results in a near $91\%$ reduction in hallucination counts compared to the original model.
The original model generated 701 and 666 hallucinations on the Near- and Far-OoD datasets, respectively, which reduced to 80 and 47 after applying our method.
Similarly, for the PASCAL-VOC task, hallucinations decrease by $85.8\%$ and $86.4\%$ for Near-OoD and Far-OoD, respectively.

\begin{table*}[ht]
    \centering
     \caption{Comparison of OoD filter performance on fine-tuned models using FPR95.}
    \label{tab:OoDeffectiveness}
    \resizebox{0.96\textwidth}{!}{%
    \renewcommand{\arraystretch}{1.2}
    \begin{tabular}{llcccccccc}
        \hline
        \textbf{ID Dataset} & \textbf{OoD Dataset} & \textbf{Condition} & \textbf{MSP}~\cite{hendrycks2017baseline} & \textbf{EBO}~\cite{liu2020energy} & \textbf{MLS}~\cite{hendrycks2022scaling} & \textbf{MDS}~\cite{lee2018simple} & \textbf{SCALE}~\cite{xu2024scaling} & \textbf{BAM}~\cite{wu2024bam} & \textbf{KNN}~\cite{sun2022out} \\
        \hline
        \multirow{4}{*}{PASCAL-VOC} 
            & \multirow{2}{*}{Near-OoD} & w/ outliers & 0.7607 & 0.9141 & 0.9110 & 0.6656 & 0.8313 & \textbf{0.4877} & 0.5675 \\
            &                           & w/o outliers & 0.6748 & 0.9049 & 0.8988 & 0.5767 & 0.8067 & \textbf{0.3804} & 0.4110 \\
            \cline{2-10}
            & \multirow{2}{*}{Far-OoD}  & w/ outliers & 0.7634 & 0.9084 & 0.9084 & 0.7863 & 0.8244 & 0.6260 & \textbf{0.6031} \\
            &                           & w/o outliers & 0.6718 & 0.9084 & 0.9008 & 0.6947 & 0.8092 & 0.5115 & \textbf{0.4580} \\
        \hline
        \multirow{4}{*}{BDD-100K} 
            & \multirow{2}{*}{Near-OoD} & w/ outliers & 0.8722 & 0.8797 & 0.8647 & 0.7594 & 0.8045 & 0.7820 & \textbf{0.6992} \\
            &                           & w/o outliers & 0.7293 & 0.8722 & 0.8647 & 0.6842 & 0.7744 & 0.7368 & \textbf{0.6015} \\
            \cline{2-10}
            & \multirow{2}{*}{Far-OoD}  & w/ outliers & 0.8118 & 0.8706 & 0.8706 & 0.9176 & 0.8000 & 0.7529 & \textbf{0.6706} \\
            &                           & w/o outliers & 0.7412 & 0.8706 & 0.8706 & 0.8235 & 0.7059 & 0.6706 & \textbf{0.5529} \\
        \hline
    \end{tabular}
    }
\end{table*}

\subsection{Generalization to Two-stage Object Detectors}
To assess  the generalizability of our approach across various model architectures, we expand our investigation beyond YOLO-based models.
Specifically, we perform comprehensive ablation studies on Faster R-CNN~\cite{ren2015faster}, a representative two-stage object detection architecture. This will help us determine whether the improvements we previously observed can also be achieved within this fundamentally different design framework.

\subsubsection{Implementation of Objectness in Faster R-CNN}
Faster R-CNN incorporates objectness estimation through a two-stage pipeline.
In the first stage, a Region Proposal Network (RPN) generates candidate regions by assigning objectness scores to a dense set of anchors. 
These scores indicate the likelihood that an anchor contains an object, helping to filter background regions before further processing. 
Based on these scores, the top-k proposals are selected for the second stage, ranked by their relative magnitude instead of absolute thresholds.
While effective in maximizing recall, the RPN's objectness scores primarily rely on  low-level visual features and are not explicitly optimized for precision~\cite{ren2015faster}.
Consequently, the RPN may produce high-scoring proposals for OoD samples, which limits its effectiveness in directly mitigating hallucinations  at the proposal stage.
The second-stage classifier processes these proposals, assigning a probability for each class, including a probability for the background class.
While background scores are computed during inference, standard Faster R-CNN workflows typically ignore them and focus only on the most probable foreground class.
However, this background confidence, denoted as $\mathit{bg\_score}$, can be utilized during inference to suppress overconfident predictions.

In this work, we adopt a modified inference rule: if the \(\mathit{bg\_score}\) is the highest among all classes, then the corresponding proposal will be discarded and not retained as a final detection.
To ensure that this modification does not compromise the model's performance on ID data, we perform a controlled evaluation using mAP as the primary metric.
The results, presented in Table~\ref{tab:impact_id_performance}, show only a slight decrease, approximately 1\%, which confirms that the adjustment maintains the model's original detection capabilities.

\begin{table}[ht]
    \centering
    \caption{Comparison of Faster R-CNN performance under different inference settings on VOC and BDD datasets.}
    \label{tab:impact_id_performance}
    \resizebox{\columnwidth}{!}{
    \renewcommand{\arraystretch}{1.2}
    \begin{tabular}{ccccccc}
        \hline
        \textbf{Dataset} & \textbf{\makecell{Inference\\ Setting}} & \textbf{mAP} & \textbf{Accuracy} & \textbf{Precision} & \textbf{Recall} & \textbf{F-score} \\
        \hline
        \multirow{2}{*}{PASCAL-VOC} 
            & w/o \(\mathit{bg\_score}\)         & 0.402 & 0.4402 & 0.5444 & 0.6968 & 0.6113 \\
            & w/ \(\mathit{bg\_score}\)    & 0.391 & 0.5034 & 0.6950 & 0.6462 & 0.6697 \\
        \hline
        \multirow{2}{*}{BDD-100K} 
            & w/o \(\mathit{bg\_score}\)          & 0.264 & 0.4929 & 0.6130 & 0.7155 & 0.6603 \\
            & w/ \(\mathit{bg\_score}\)     & 0.252 & 0.5324 & 0.7617 & 0.6389 & 0.6949 \\
        \hline
    \end{tabular}
    }
\end{table}

\begin{figure*}[t]
    \centering
    \begin{minipage}[b]{0.45\textwidth}
        \centering
        \includegraphics[width=0.9\linewidth, height=5cm]{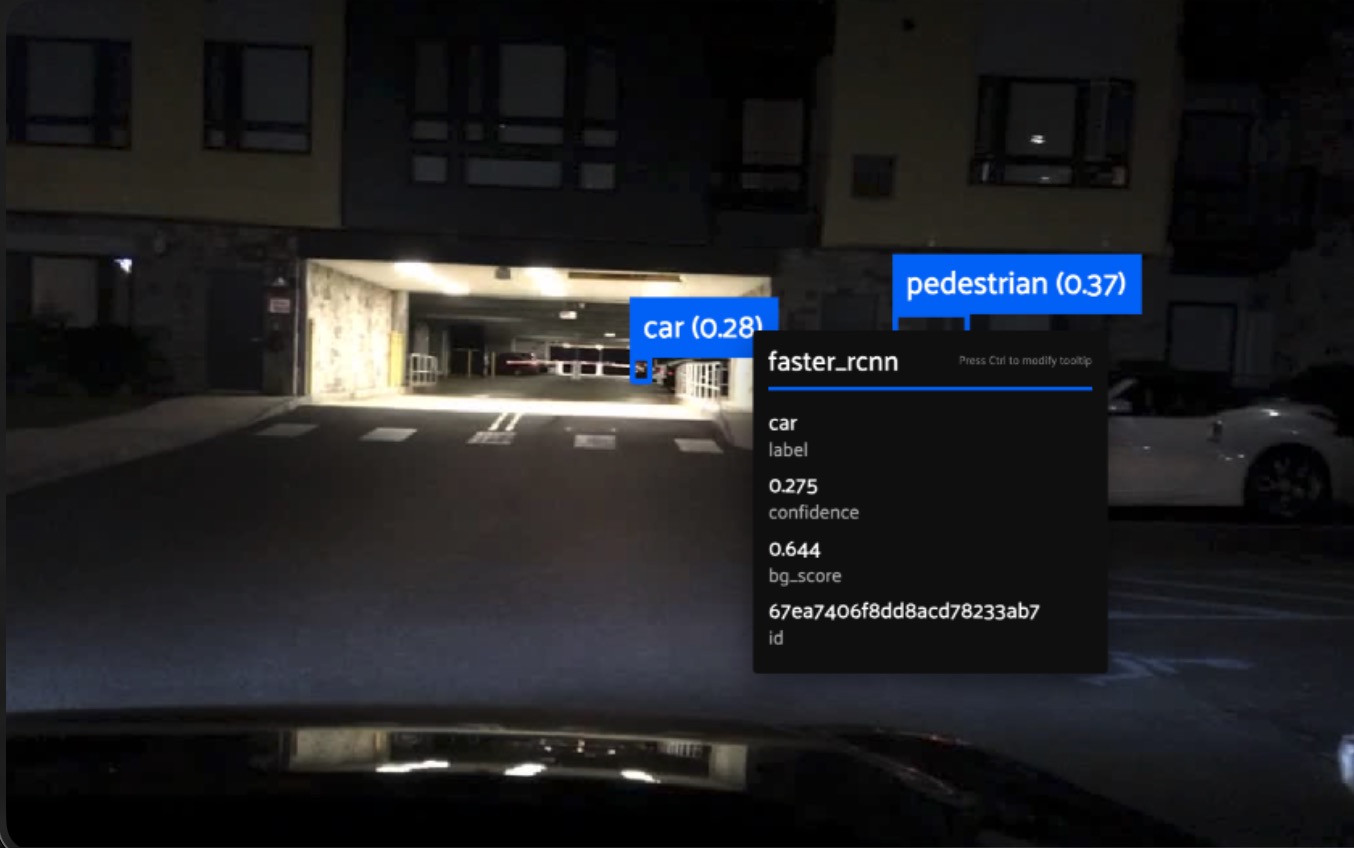}
    \end{minipage}
    \hspace{3em}
    \begin{minipage}[b]{0.45\textwidth}
        \centering
        \includegraphics[width=0.9\linewidth, height=5cm]{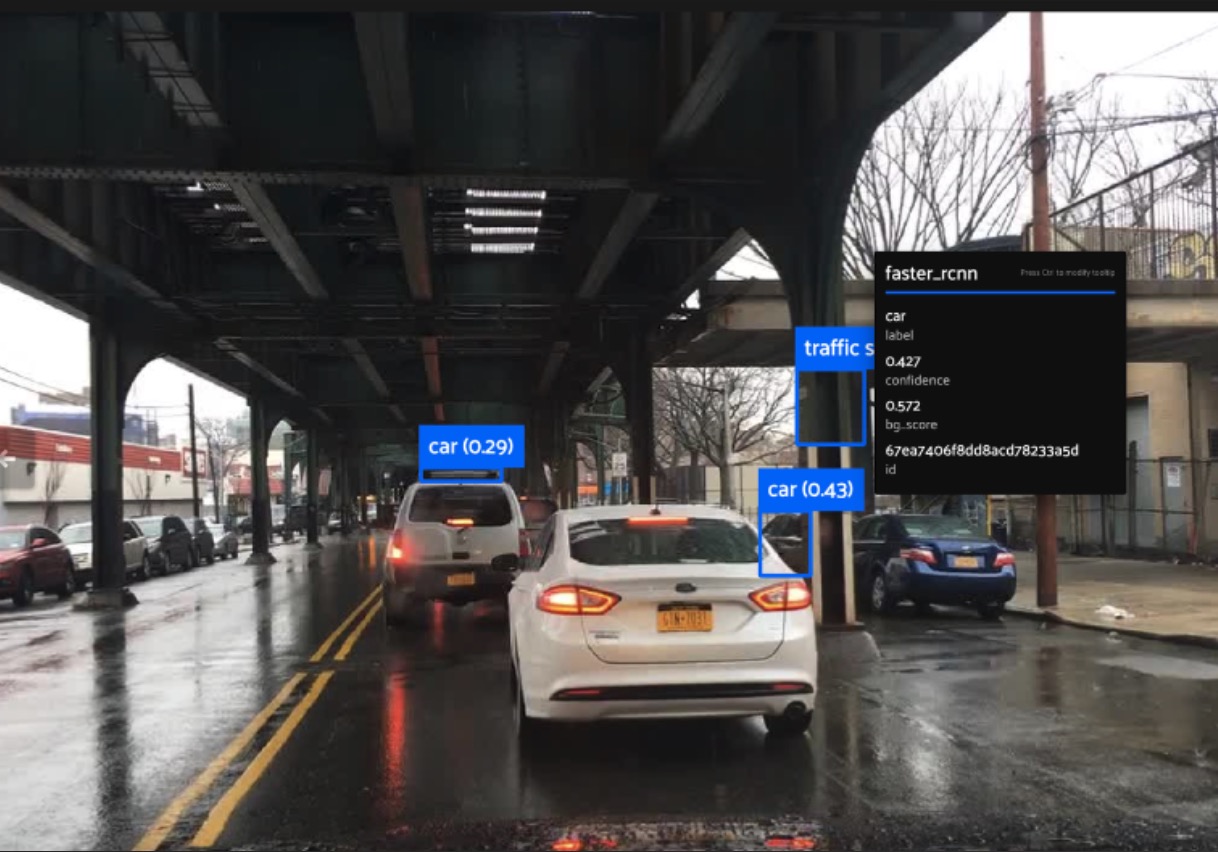}
    \end{minipage}
    \caption{Visualization of detection instances in BDD-100K with higher background score than confidence score. Left: a distant, partially hidden car in a peripheral region; Right: several small or partially occluded objects under difficult lighting. The space between the images is adjustable for optimal layout. Both images are shown at equal size for consistency.}
    \label{fig:quali_bg}
\end{figure*}

We also conducted a qualitative analysis of filtered detections to demonstrate how the background-score threshold affects different types of predictions.
The results indicate that most filtered predictions appeared in peripheral image regions, involved small or partially occluded objects, or occurred under difficult lighting conditions.
Fig.~\ref{fig:quali_bg} shows cases where the background score surpassed the confidence of the predicted foreground class.
The left image shows a distant, hidden car, while the right image features several small and partially visible objects.
The filtered detections were mostly false positives with low IoU, which accounts for the minimal effect  on mAP.
While  a few true positives were incorrectly removed, the resulting subsequent drop in recall was balanced by a greater increase in precision, as demonstrated in Table~\ref{tab:impact_id_performance}.

\subsubsection{Fine-tuning Strategy for Faster R-CNN}
\begin{figure}[t]
    \centering
    \includegraphics[width=\columnwidth]{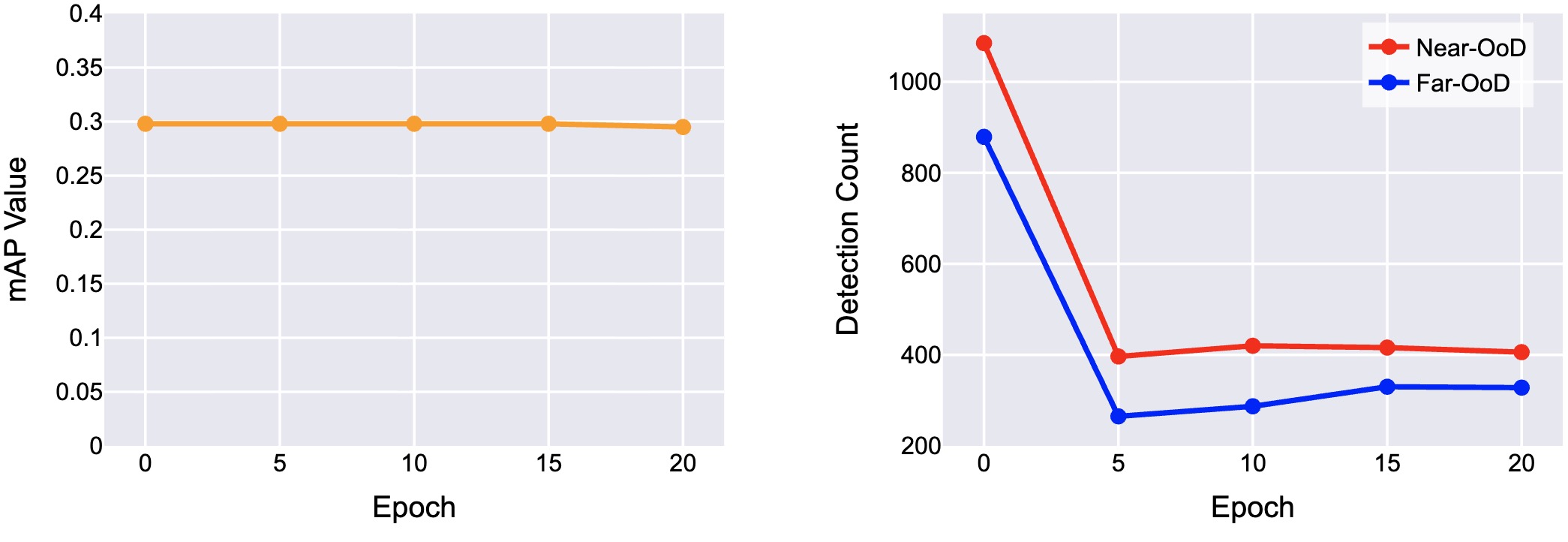}
    \caption{Fine-tuning trends of Faster-RCNN model pre-trained on BDD-100K: mAP evolution (left), and prediction counts on OoD datasets (right).}
    \label{fig:fasterrcnn_trends}
\end{figure}

Building on the architectural analysis above, we fine-tune Faster R-CNN, focusing on leveraging the background class scores from the second-stage classifier.
During fine-tuning, we freeze the backbone and the Feature Pyramid Network (FPN), and make only the Region Proposal Network (RPN) and the second-stage Region of Interest (RoI) head trainable.
We adopt the same loss design as in our previous YOLO-based models, with a class-conditional weighting scheme to emphasize proximal samples.
Mis-prediction of proximal samples are penalized ten times more than those on ID samples.
This weighting helps the second-stage classifier confidently assign high background scores to OoD content.
Consistent with this setup, we retain the pretrained feature extraction capabilities by freezing the backbone and FPN, updating only the RPN and the second-stage classifier (RoI head).
The models are fine-tuned for 20 epochs, with performance monitored on both proximal and test OoD datasets to assess their stability and effectiveness.
Fig.~\ref{fig:fasterrcnn_trends} illustrates the fine-tuning dynamics, showing the evolution of mAP on ID data (left), and the number of predictions on OoD datasets (right) over training epochs.
These trends indicate that our approach maintains task performance on ID samples while substantially reducing hallucinations on OoD samples within just a few epochs.

\subsubsection{Empirical Results and Analysis}
To evaluate the effectiveness of our method on Faster R-CNN, we examine the decrease in hallucinations before and after fine-tuning under the same conditions as our experiments with YOLO models, as illustrated in Table~\ref{tab:ood_counts_fasterrcnn}.
The fine-tuning process demonstrates substantial reductions in hallucination across all OoD datasets. 
In the BDD-100K dataset, the number of hallucination count over Near-OoD samples drops from 2576 to 394, indicating an 84.7\% reduction.
Similarly, Far-OoD hallucination counts decrease from 1634 to 242, showing an 85.2\% reduction.

Similar to YOLO-based models, we also examine the effect of combining fine-tuning with OoD detection for Faster R-CNN.
We start by evaluating the effectiveness of various OoD detection methods in Table~\ref{tab:OoDeffectiveness_FasterRCNN} and choose the best-performing one for integration.
The results show that KNN consistently achieves the best performance across most dataset combinations.
By integrating the KNN OoD detector with the fine-tuned Faster R-CNN model, our approach achieves further substantial reductions in hallucinations, as shown in Table~ \ref{tab:ood_counts_fasterrcnn}.
For the BDD-100K dataset, the combination of the fine-tuned model and the KNN OoD filter led to a significant reduction in Near-OoD hallucinations, decreasing from 2576 to just 207 (a reduction of 92.0\%).
Similarly, Far-OoD hallucinations decreased from 1634 to 167, representing a reduction of 89.8\%.
On the PASCAL-VOC benchmark, Near-OoD hallucinations decreased from 2150 to 710 (a reduction of 67.0\%), while Far-OoD hallucinations dropped from 1335 to 253 (a reduction of 81.0\%).
This ablation study demonstrates that our approach can be effectively adapted to different object detection frameworks, provided there are some architecture-specific modifications.
This confirms the fundamental validity and generalizability of our approach.

\begin{table}[t]
\caption{OoD counts comparison of different Faster-RCNN models.}
    \label{tab:ood_counts_fasterrcnn}
    \centering
    \renewcommand{\arraystretch}{1.2}
    \resizebox{\columnwidth}{!}{ 
\begin{tabular}{cccccc}
    \hline
    \textbf{\makecell{ID Dataset}} & \textbf{\makecell{OoD Datasets}} & \textbf{Original} & \textbf{\makecell{Original+KNN}} & \textbf{Fine-tuned} & \textbf{\makecell{Fine-tuned+KNN}} \\
    \hline
    \multirow{2}{*}{PASCAL-VOC} 
        & Near-OoD & 2150 & 1332 & 1201 & \textbf{710} \\
        & Far-OoD & 1335 & 501 & 596 & \textbf{253} \\
    \midrule
    \multirow{2}{*}{BDD-100K} 
        & Near-OoD & 2576 & 1302 & 394 & \textbf{207} \\
        & Far-OoD & 1634 & 813 & 242 & \textbf{167} \\
    \hline
\end{tabular}
}
\end{table}

\begin{table*}[t]
    \centering
    \caption{Comparison of OoD filter performance on fine-tuned Faster-RCNN models using FPR95 .}
    \label{tab:OoDeffectiveness_FasterRCNN}
    \resizebox{0.85\textwidth}{!}{%
    \renewcommand{\arraystretch}{1.2}
    \begin{tabular}{ccccccccc}
        \hline
        \textbf{ID Dataset} & \textbf{OoD Dataset} & \textbf{MSP}~\cite{hendrycks2017baseline} & \textbf{EBO}~\cite{liu2020energy} & \textbf{MLS}~\cite{hendrycks2022scaling} & \textbf{MDS}~\cite{lee2018simple} & \textbf{SCALE}~\cite{xu2024scaling} & \textbf{BAM}~\cite{wu2024bam} & \textbf{KNN}~\cite{sun2022out} \\
        \hline
        \multirow{2}{*}{PASCAL-VOC} 
            & Near-OoD & 0.6836 & 0.6062 & 0.5962 & 0.4996 & 0.9234 & 0.6295 & \textbf{0.5912} \\
            & Far-OoD  & 0.7869 & 0.5621 & 0.5789 & 0.5638 & 0.8070 & 0.4698 & \textbf{0.4245} \\
        \hline
        \multirow{2}{*}{BDD-100K} 
            & Near-OoD & 0.7746 & 0.9483 & 0.9255 & 0.7890 & 0.8635 & 0.7384 & \textbf{0.5429} \\
            & Far-OoD  & 0.7441 & 0.9437 & 0.9108 & 0.7934 & 0.8498 & 0.7207 & \textbf{0.5634} \\
        \hline
    \end{tabular}
    }
\end{table*}

\subsection{Generalization to Transformer-based Object Detectors}
To further evaluate the generalizability of our approach beyond CNN-based architectures, we extend our study to recent Transformer-based object detectors.
Specifically, we experiment with RT-DETR~\cite{zhao2024detrs}, a real-time end-to-end model that integrates a Transformer-based decoder with a convolutional backbone, achieving high accuracy with efficient inference.
RT-DETR employs a sparse, query-based detection strategy with \textit{implicit objectness scoring} penalizing high-confidence predictions on background regions similar to YOLOv10.
We therefore reuse the same fine-tuning strategy to mitigate hallucinations on Transformer-based architectures.

To assess the impact of our fine-tuning method on RT-DETR, we compare hallucination counts before and after fine-tuning under identical conditions (see Table~\ref{tab:ood_counts_rtdetr}).
Across all OoD datasets, fine-tuning yields substantial reductions in hallucinated detections.
On the BDD-100K dataset, hallucinations on Near-OoD samples drop from 3145 to 967 (a reduction of 69.3\%), while Far-OoD samples show a decrease from 1220 to 426 (a reduction of 65.1\%).
For the PASCAL-VOC task, we observe similar improvements: Near-OoD hallucinations are reduced from 2311 to 964 (a reduction of 58.3\%), and Far-OoD hallucinations from 1589 to 966 (a reduction of 39.2\%).

Similar to our previous setups, we evaluate the effect of combining fine-tuning with OoD detection for RT-DETR by leveraging raw classification logits.
We begin by assessing several OoD detection methods and select the best-performing one for integration.
Results in Table~\ref{tab:oodDETR} show that both KNN and BAM consistently outperform other baselines cross the evaluated settings, with KNN achieving the highest accuracy in the majority of cases.
By integrating the KNN OoD detector with the fine-tuned RT-DETR model, our approach achieves further substantial reductions in hallucinations, as shown in Table~\ref{tab:ood_counts_rtdetr}. 
Integrating KNN with the fine-tuned RT-DETR yields further substantial reductions in hallucinations (see Table~\ref{tab:ood_counts_rtdetr}).
On BDD-100K, hallucinations on Near-OoD samples drop from 3145 to 525 (83.3\% reduction), and on Far-OoD from 1220 to 240 (80.3\% reduction).
On PASCAL-VOC, the combined approach achieves similar gains, reducing hallucinations by 83.3\% on Near-OoD and 70.4\% on Far-OoD samples.
These results demonstrate that our method can be effectively extended to Transformer-based detection frameworks, reinforcing its generality across architectural paradigms.

\begin{table}[h]
    \caption{OoD counts comparison of RT-DETR models.}
        \label{tab:ood_counts_rtdetr}
        \centering
        \renewcommand{\arraystretch}{1.2}
        \resizebox{0.48\textwidth}{!}{ 
    \begin{tabular}{cccccc}
        \hline
        \textbf{\makecell{ID \\ Dataset}} & \textbf{\makecell{OoD \\ Datasets}} & \textbf{Original} & \textbf{\makecell{Original+KNN}} & \textbf{Fine-tuned} & \textbf{\makecell{Fine-tuned+KNN}} \\
        \hline
        \multirow{2}{*}{PASCAL-VOC}
            & Near-OoD & 2311 & 1643 & 964 & \textbf{386} \\
            & Far-OoD & 1589 & 1001 & 966 & \textbf{470} \\
        \midrule
        \multirow{2}{*}{BDD-100K} 
            & Near-OoD & 3145 & 1952 & 967 & \textbf{525} \\
            & Far-OoD & 1220 & 708 & 426 & \textbf{240} \\
        \hline
    \end{tabular}
    }
\end{table}

\begin{table*}[t]
    \centering
    \caption{Comparison of OoD filter performance on fine-tuned RT-DETR models using FPR95.}
    \label{tab:oodDETR}
    \resizebox{0.85\textwidth}{!}{%
    \renewcommand{\arraystretch}{1.2}
    \begin{tabular}{ccccccccc}
        \hline
        \textbf{ID Dataset} & \textbf{OoD Dataset} & \textbf{MSP}~\cite{hendrycks2017baseline} & \textbf{EBO}~\cite{liu2020energy} & \textbf{MLS}~\cite{hendrycks2022scaling} & \textbf{MDS}~\cite{lee2018simple} & \textbf{SCALE}~\cite{xu2024scaling} & \textbf{BAM}~\cite{wu2024bam} & \textbf{KNN}~\cite{sun2022out} \\
        \hline
        \multirow{2}{*}{PASCAL-VOC} 
                & Near-OoD & 0.7044 & 0.9803 & 0.9284 & 0.4865 & 0.8112 & \textbf{0.3880} & 0.4004 \\
                & Far-OoD  & 0.6781 & 0.9669 & 0.8975 & 0.5290 & 0.7692 & 0.4959 & \textbf{0.4865} \\
        \hline
        \multirow{2}{*}{BDD-100K} 
                & Near-OoD & 0.7746 & 0.9483 & 0.9255 & 0.7890 & 0.8635 & 0.7384 & \textbf{0.5429} \\
                & Far-OoD  & 0.7441 & 0.9437 & 0.9108 & 0.7934 & 0.8498 & 0.7207 & \textbf{0.5634} \\
        \hline
    \end{tabular}
    }
\end{table*}

\subsection{Ablation on Fine-Tuning Strategy}

Our initial experiments followed a zero-shot setup, where the model was fine-tuned on a diverse set of proximal OoD categories and tested on entirely disjoint categories, ensuring no category-level overlap between training and testing.
In this section, we explore a complementary and more practical setting: \textbf{few-shot fine-tuning using OoD data from the same category as the evaluation set}.
We refer to this as the \textit{few-shot setting}, where limited target-category data is available to refine the model after deployment, for example, when a specific failure mode (e.g., hallucinating horses as pedestrians) is discovered.

\begin{figure*}[t]
    \centering
    \includegraphics[width=0.9\textwidth]{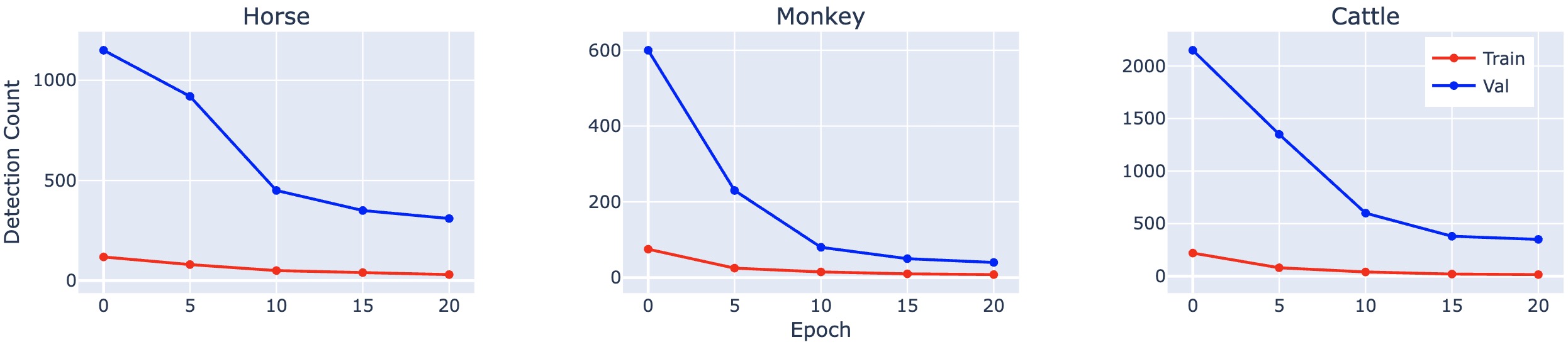}
    \caption{Evolution of pedestrian detection counts during few-shot learning across different proximal OoD animal categories. Each subplot shows training curves (10\% of data) and validation curves (90\% of data) for horses, monkeys, and cattle. All categories demonstrate substantial and consistent reductions in pedestrian hallucinations, with training curves approaching zero while validation improvements plateau after approximately 15 epochs.}
    \label{fig:fsl_ablation_ood_categories}
\end{figure*}

\textbf{Effectiveness Across Categories.}  
We conducted experiments on several OoD categories known to trigger hallucinations in BDD-trained models, including \textit{horses}, \textit{monkeys}, and \textit{cattle}.
For each category, we curated the dataset using our data curation pipeline (see Sec.~\ref{sec:pipeline}), leveraging OpenImages V7 as the data source to ensure high-quality, category-specific images.
Specifically, we collected around 1,000 images per category containing only OoD objects, using 10\% (100 images) for fine-tuning and reserving the remaining 90\% for evaluation.
This setting simulates a realistic few-shot adaptation scenario.
Across all three categories, we observed substantial reductions in hallucinated detections as shown in Fig.~\ref{fig:fsl_ablation_ood_categories}.
For instance, pedestrian false positives on horse images dropped from over 1,000 to roughly 300 after fine-tuning.
Similar trends were observed for monkeys and cattle, with hallucination reductions exceeding 70\% in each case.
These results confirm that our fine-tuning strategy remains highly effective even when applied to small amounts of same-category data, and that it generalizes well across diverse hallucination sources.

\begin{figure}[t]
    \centering
    \includegraphics[width=0.7\columnwidth]{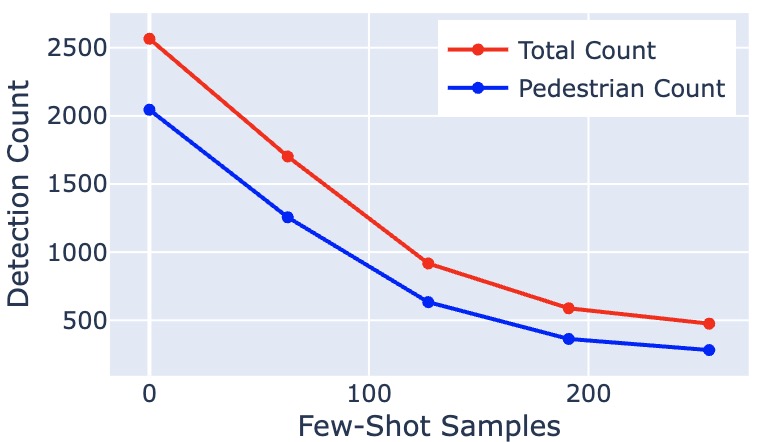}
    \caption{Impact of few-shot training sample size on hallucination reduction for horse OoD test images. Both total detections and pedestrian-specific misclassifications decrease substantially with increasing sample size, demonstrating strong data efficiency with diminishing returns beyond 200 samples.}
\label{fig:fsl_ablation_sample_counts}
\end{figure}

\textbf{Impact of Sample Quantity.}  
We further studied how the number of fine-tuning samples affects hallucination suppression.
Using the horse dataset as a case study, we varied the training set size from 0 to 256 images and measured hallucination counts on a held-out test set of 1,500 horse images.
The results in~\ref{fig:fsl_ablation_sample_counts} show a steep initial gain: with just 64 examples, hallucinations decreased by more than 60\%, and with 200 examples, the reduction exceeded 80\%.
However, the marginal benefit diminishes beyond 200 samples, suggesting that most of the gain can be achieved with modest data.
Notably, pedestrian-specific misclassifications dominate the hallucination count and benefit most from the few-shot correction.

These findings highlight the practicality and flexibility of our approach.
While the original zero-shot setting demonstrates the generality of hallucination suppression, the few-shot variant offers an efficient mechanism for addressing category-specific failure cases using minimal data.
This allows for fast, targeted model refinement in deployment, with strong performance gains even under tight data constraints.

\subsection{Ablation on Fine-Tuning Data Choices}

We analyze the sensitivity of our method to fine-tuning data along three dimensions: (i) the selection of ID fine-tuning data, (ii) the proximal OoD data source, and (iii) the proximal OoD selection metric.

\paragraph{Sensitivity to ID Fine-Tuning Data} We construct an error-prone ID subset (denoted as ID$_{\text{err}}$) by ranking images according to the number of erroneous predictions produced by the baseline model and selecting the top 2,000 images.
Inspired by \textit{replay} in continual learning~\cite{rolnick2019experience}, this subset enables effective fine-tuning using substantially less ID data.
As shown in Table~\ref{tab:ablation_proximal_source}, the effectiveness of our method is largely preserved under this setting, suggesting that fine-tuning on a small ID$_{\text{err}}$ subset can maintain ID performance close to that of the whole ID set (ID$_{\text{all}}$), while suppressing hallucinations.

\begin{table}[ht]
\centering
\small
\caption{Sensitivity to ID data choice and proximal OoD data source.}
\label{tab:ablation_proximal_source}
\setlength{\tabcolsep}{6pt}
\renewcommand{\arraystretch}{1.12}
\resizebox{0.9\columnwidth}{!}{%
\begin{tabular}{cccc}
\hline
\textbf{ID\,/\,OoD setting} & \textbf{mAP} $\uparrow$ & \textbf{near-OoD det.} \ $\downarrow$ & \textbf{far-OoD det.}\ $\downarrow$ \\
\hline
ID$_{\text{all}}$, Caltech256 & \textbf{0.2556} & 133 & 85 \\
ID$_{\text{err}}$, Caltech256 & 0.2492 & 170 & 63 \\
ID$_{\text{err}}$, Object365 & 0.2428 & \textbf{118} & \textbf{55} \\
\hline
\end{tabular}%
}
\end{table}

\begin{table}[h]
\centering
\small
\caption{Sensitivity to semantic distance of proximal OoD data. Closer proximal OoD samples reduce OoD detection counts but slightly lower ID performance (mAP-all\,/\,small\,/\,occluded), while more distant samples preserve ID performance with higher OoD counts.}
\label{tab:ablation_semantic_distance}
\setlength{\tabcolsep}{3pt}
\renewcommand{\arraystretch}{1.12}
\resizebox{\columnwidth}{!}{%
\begin{tabular}{lccc}
\hline
\textbf{Setting} & \textbf{\makecell{mAP \\ all\,/\,small\,/\,occluded}} $\uparrow$ & \textbf{near-OoD det.}\ $\downarrow$ & \textbf{far-OoD det.}\ $\downarrow$ \\
\hline
Vanilla & \textbf{0.2628}\,/\,\textbf{0.0278}\,/\,\textbf{0.0661} & 770 & 574 \\
LLM-recommended & 0.2506\,/\,0.0259\,/\,0.0649 & 219 & 93 \\
Text embedding (top-3) & 0.2457\,/\,0.0261\,/\,0.0622 & \textbf{143} & \textbf{60} \\
Text embedding (bottom-3) & 0.2523\,/\,0.0257\,/\,0.0653 & 226 & 108 \\
Visual embedding (top-3) & 0.2504\,/\,0.0260\,/\,0.0652 & 233 & 89 \\
Visual embedding (bottom-3) & 0.2530\,/\,0.0260\,/\,0.0659 & 239 & 132 \\
\hline
\end{tabular}%
}
\end{table}

\paragraph{Proximal OoD Data Source} We evaluate sensitivity to the choice of proximal OoD dataset by comparing Caltech256, used in the main experiments, with Object365~\cite{shao2019objects365}, while keeping the category set and training protocol fixed.
The results in Table~\ref{tab:ablation_proximal_source} show that replacing Caltech256 with Object365 further reduces OoD detections, particularly for near-OoD cases, while only slightly affecting ID performance, indicating that our method is robust to the choice of proximal OoD dataset.

\paragraph{Proximal Data Selection Metric}
\label{sec:ablation_proximal_metric}
We further examine the effect of the metric used to select proximal OoD samples.
In addition to the LLM-based expert recommendations used in the main experiments, we consider ranking candidate samples using text or visual embeddings from CLIP~\cite{radford2021learning}.
All other training settings are kept identical.
Specifically, we rank proxy categories by cosine similarity computed from either CLIP text embeddings of category names or CLIP visual embeddings of representative images, and select the top-3 (closest) or bottom-3 (farthest) categories per ID category. 
For fine-tuning, we sample 500 images from the selected proxy categories, along with ID$_\text{err}$ dataset.
Table~\ref{tab:ablation_semantic_distance} shows that our method behaves as expected: selecting semantically closer proximal OoD samples (higher similarity) yields most significant reduction in both far- and near-OoD hallucinated detections, but slightly reduces ID mAP; selecting more distant proximal OoD samples (lowest similarity) better preserves ID performance at the cost of higher residual OoD detections.
In addition, performance on hard ID modes, including small and occluded objects, remains comparable under the closest proximal OoD settings.

\subsection{Computational Efficiency Analysis}

To evaluate the practical deployment viability of our approach, we analyze the computational overhead from two complementary perspectives: (i) the fine-tuned detector itself, and (ii) the external filtering component.
Since our fine-tuning strategy involves no architectural modifications to the object detection model, the runtime and memory performance remain identical to the original baseline during inference.
When an external KNN-based OoD filtering module is additionally integrated for enhanced hallucination reduction, it incurs only minimal overhead.
Specifically, the filtering component introduces an additional latency of 0.30\,ms, throughput of 3,375.9 predictions per second, and an additional memory footprint of only 4.0\,MB.
These characteristics demonstrate that our complete pipeline maintains excellent computational efficiency, making it practical for deployment in real-time and resource-constrained environments.

%% file: sections/6_in-depth_analysis.tex
\section{WHY IT WORKS: AN IN-DEPTH ANALYSIS}\label{sec:analysis}
To understand why our fine-tuning strategy on proximal OoD data effectively suppresses hallucinations, we provide a deeper analysis from two complementary perspectives: attention shifts in the model’s spatial focus, and changes in its confidence behavior under OoD conditions.

\subsection{XAI-based Interpretation}
In this subsection, we apply saliency map-based XAI techniques to analyze why fine-tuning with proximal OoD data effectively reduces hallucinations.
Specifically, we visualize where the model focuses its attention before and after fine-tuning on both ID and OoD inputs.
To generate object-specific explanations for YOLOv10, we design a feature extractor inspired by the RoIAlign operation in Mask R-CNN~\cite{he2018maskrcnn}, enabling precise localization of discriminative regions within detected objects.
We extract intermediate feature maps from the multi-scale Feature Pyramid Network (FPN) to capture object representations at different spatial resolutions.
For each detected object, RoIAlign precisely crops and aligns the relevant features from all pyramid levels based on its bounding box.
Unlike traditional RoI pooling, RoIAlign avoids quantization artifacts by using bilinear interpolation, ensuring high spatial fidelity.
The aligned features are then average-pooled to produce a compact 2D activation map, where red/orange regions indicate strong feature contributions and blue/green regions indicate weaker ones.

%

\begin{figure}[ht] 
    \centering
    \includegraphics[width=0.92\columnwidth]{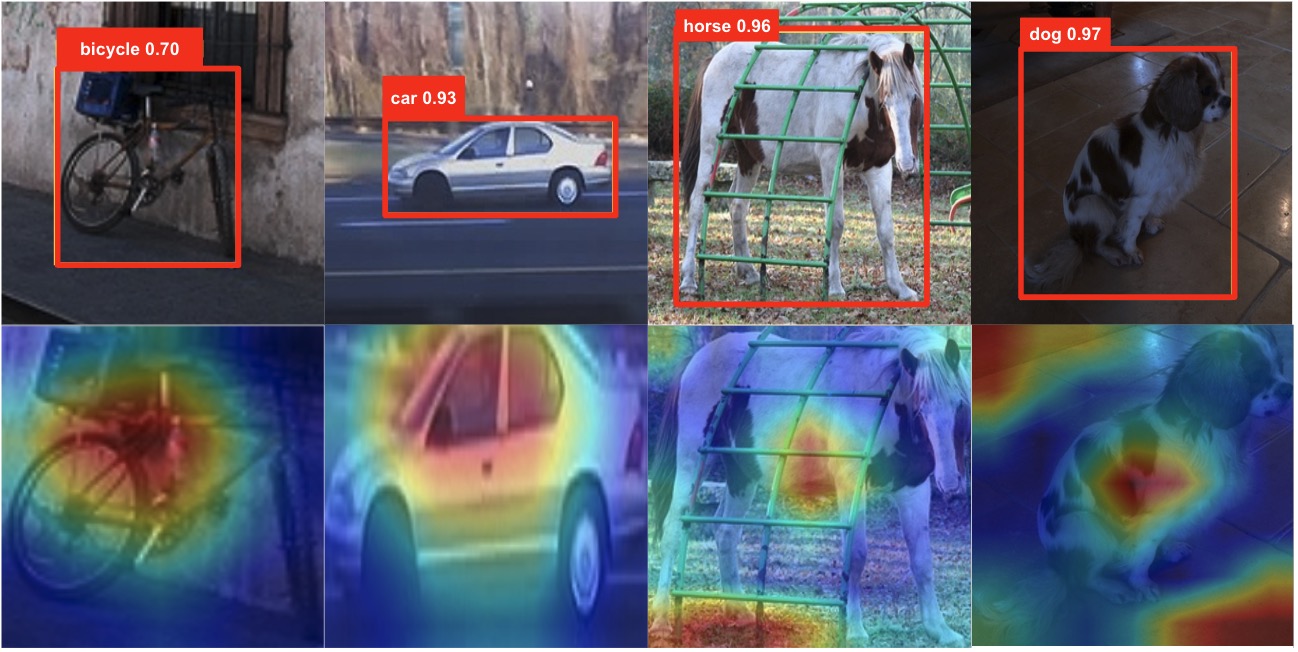} 
    \caption{Saliency map visualization for ID samples on PASCAL-VOC dataset showing correct predictions by the vanilla YOLOv10 model. The top row displays original images with detection results, while the bottom row shows corresponding saliency maps highlighting regions of high neural activation. The model focuses primarily on object features and relevant contextual information for accurate classification.}
    \label{fig:xai-id}
\end{figure}

\begin{figure}[t] 
    \centering
    \includegraphics[width=0.92\columnwidth]{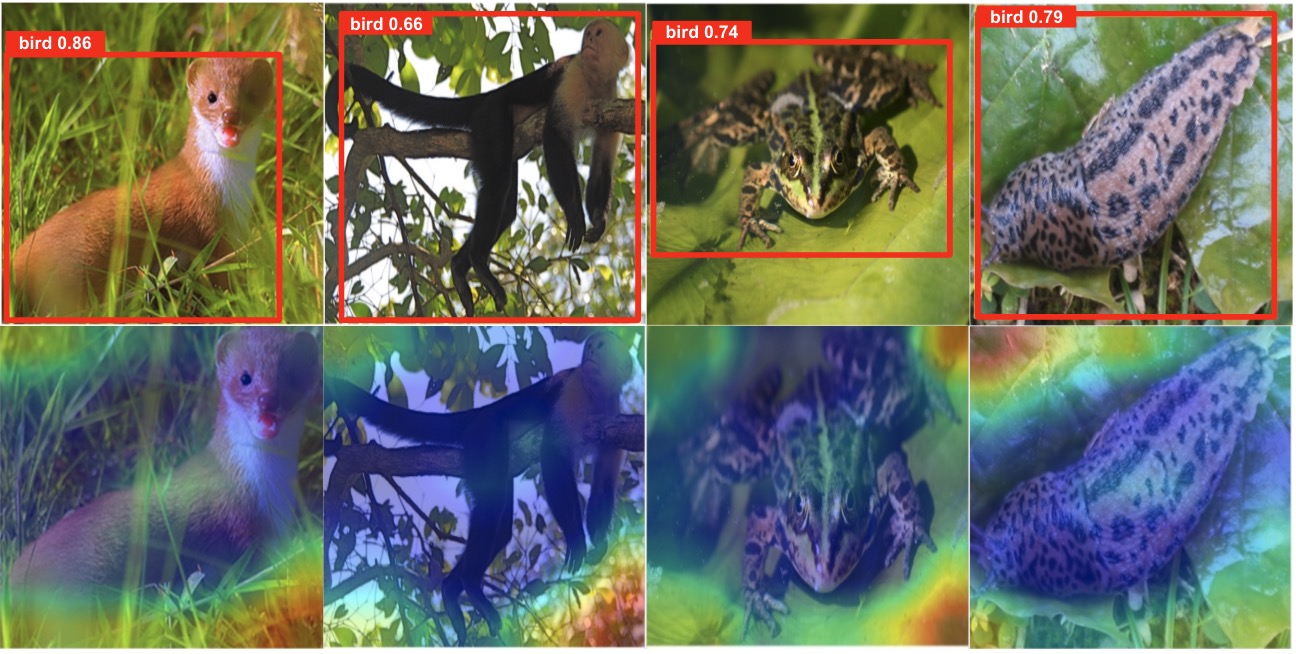} 
    \caption{Saliency map visualization for OoD samples on Near-OoD dataset showing hallucinated predictions by the vanilla YOLOv10 model. The top row displays original images with erroneous bird detections, while the bottom row shows corresponding saliency maps. The model inappropriately focuses on background regions and edges rather than actual object features, revealing the spurious correlations that lead to hallucinations.}
    \label{fig:xai-ood-hallucinations}
\end{figure}

\begin{figure}[t] 
    \centering
    \includegraphics[width=\columnwidth]{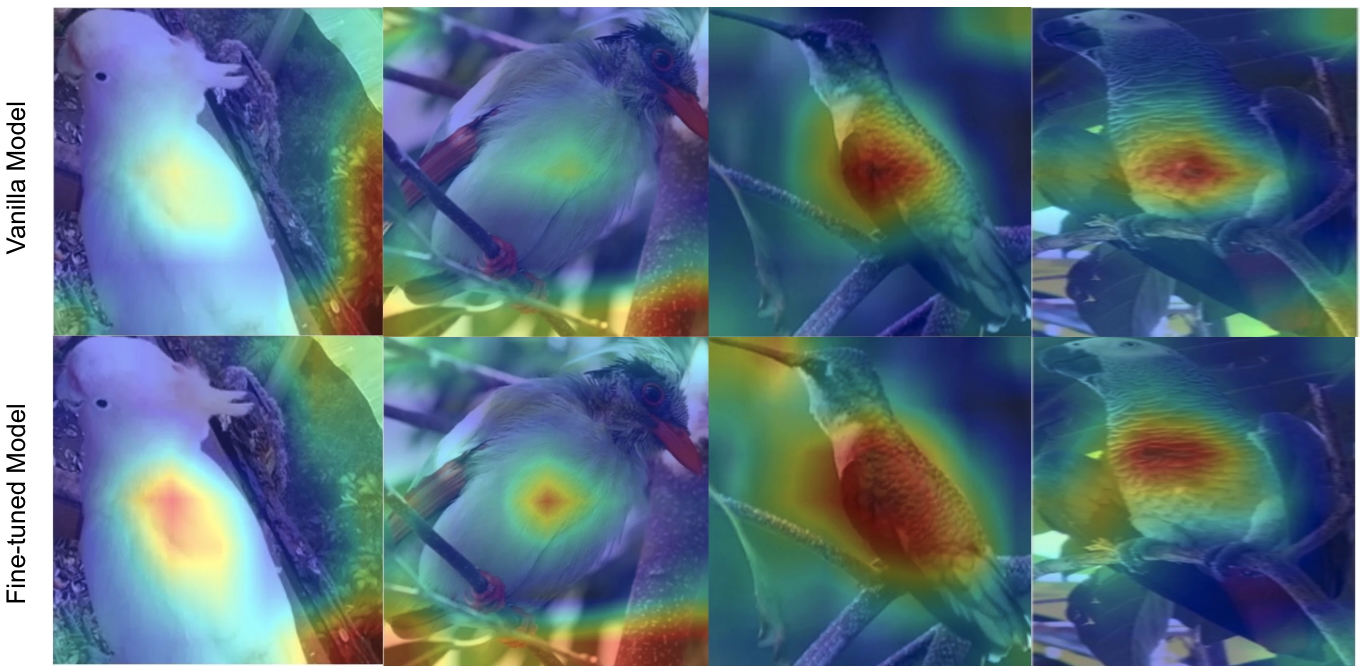} 
    \caption{Saliency map comparison between vanilla (top row) and fine-tuned (bottom row) YOLOv10 models on ID samples on PASCAL-VOC dataset. The fine-tuned model demonstrates improved focus on object-relevant features with reduced background artifacts, while the vanilla model shows spurious activations in background regions. Red/orange areas indicate high activation levels that contribute most to the model's decisions.}
    \label{fig:xai-finetune}
\end{figure}

We use the PASCAL-VOC dataset, which contains larger objects, for clearer saliency map analysis, since objects in BDD-100K are often small and difficult to visualize.
We compare model attention on ID versus OoD inputs to reveal how hallucinations emerge.
As shown in Fig.\ref{fig:xai-id}, saliency maps on ID images highlight the core body of each object, occasionally extending to nearby context.
This focus suggests the model relies primarily on relevant object features for accurate predictions.
In contrast, Fig.\ref{fig:xai-ood-hallucinations} shows that hallucinated detections on OoD inputs are driven by attention to edges, corners, or background textures rather than the actual object.
For example, many bird hallucinations occur in green, tree-filled backgrounds, where the model fixates on the vegetation instead of object features.
This behavior is likely due to spurious correlations learned from bird instances in the ID training set.
This behavior aligns with prior work~\cite{jiang2024comparing}, which suggests that CNNs often make disjunctive decisions, triggering predictions from partial cues via OR-like logic.
In our case, the model learns to detect a class based on either object features or context.
On OoD inputs, where object features are absent but context cues remain, this disjunctive behavior causes the model to hallucinate detections based solely on misleading background signals.

As a final investigation, we compare saliency maps from the vanilla and fine-tuned YOLOv10 models to assess how fine-tuning alters the model’s decision process.
Since hallucinations largely disappear after fine-tuning, we analyze changes using ID inputs, where both models produce valid predictions.
As shown in Fig.~\ref{fig:xai-finetune}, the vanilla model often exhibits strong activations in background regions, while the fine-tuned model focuses more precisely on the object itself.
This shift suggests that fine-tuning suppresses reliance on spurious background cues, promoting tighter localization of object-relevant features.
We hypothesize that proximal OoD samples used as negative examples during fine-tuning help the model learn that background context alone is insufficient to justify a detection.
As a result, the model becomes less sensitive to misleading contextual patterns and more robust against hallucinations in OoD scenarios.

\subsection{Confidence-driven Insights}
We further investigate the effectiveness of our fine-tuning strategy by analyzing how it reshapes the model’s prediction confidence on OoD inputs.
Specifically, we examine whether training on proximal OoD data encourages the model to assign lower confidence to OoD objects, thereby reducing hallucinated detections.

\begin{figure}[t]
    \centering
    \includegraphics[width=\columnwidth]{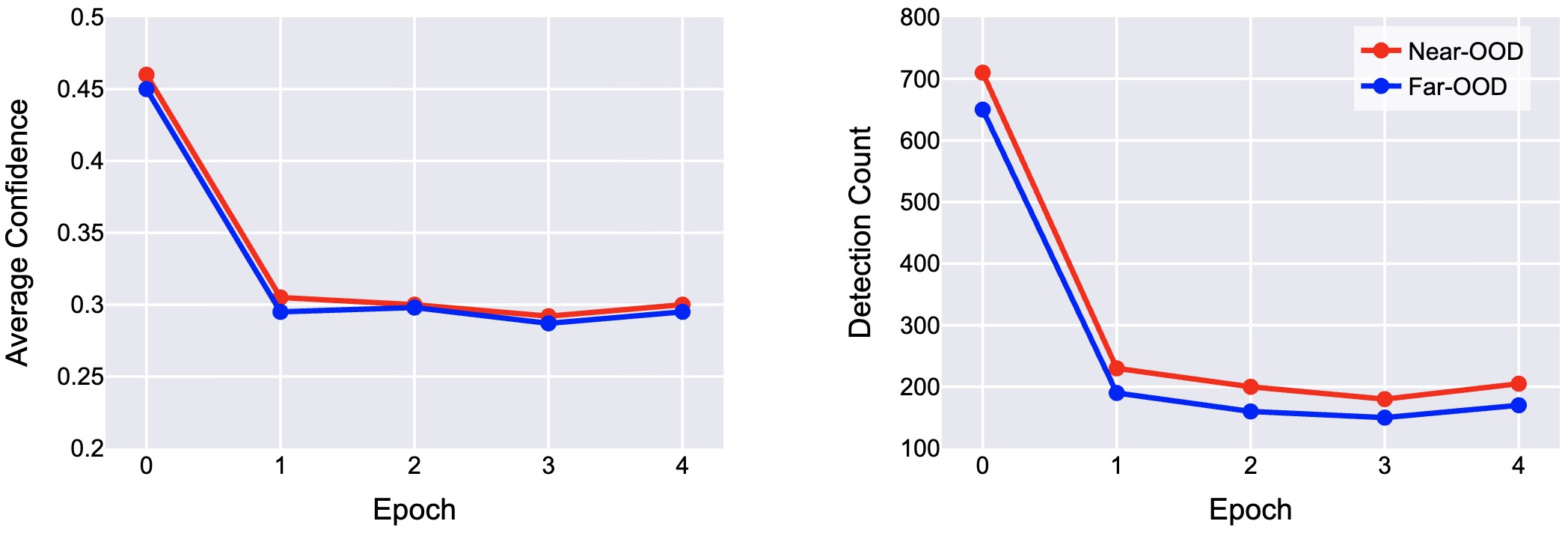} 
    \caption{Evolution of average confidence scores (left) and total detection counts (right) on Near-OoD and Far-OoD datasets across fine-tuning epochs. Both metrics demonstrate correlated declining trends, indicating that the model becomes increasingly conservative in making predictions on out-of-distribution data as fine-tuning progresses, effectively reducing hallucinations.}
    \label{fig:ood_confidence_detection_trends_epoch}
\end{figure}

To test this hypothesis, we track the average confidence scores and detection counts on both Near-OoD and Far-OoD datasets across fine-tuning epochs.
As shown in Fig.~\ref{fig:ood_confidence_detection_trends_epoch}, both metrics exhibit a clear downward trend.
Average confidence scores drop sharply after the first epoch and continue to decline gradually thereafter.
This reduction is mirrored in the total number of detections: for example, on the Near-OoD set, detections fall from roughly 700 to 200, while average confidence drops from 0.46 to 0.29.
Similar trends are observed on the Far-OoD set, where detections decrease from about 640 to 170.
These results suggest that the model becomes increasingly conservative on OoD inputs, and that confidence suppression directly contributes to hallucination mitigation.


\begin{figure*}[t] 
    \centering
    \includegraphics[width=0.95\textwidth]{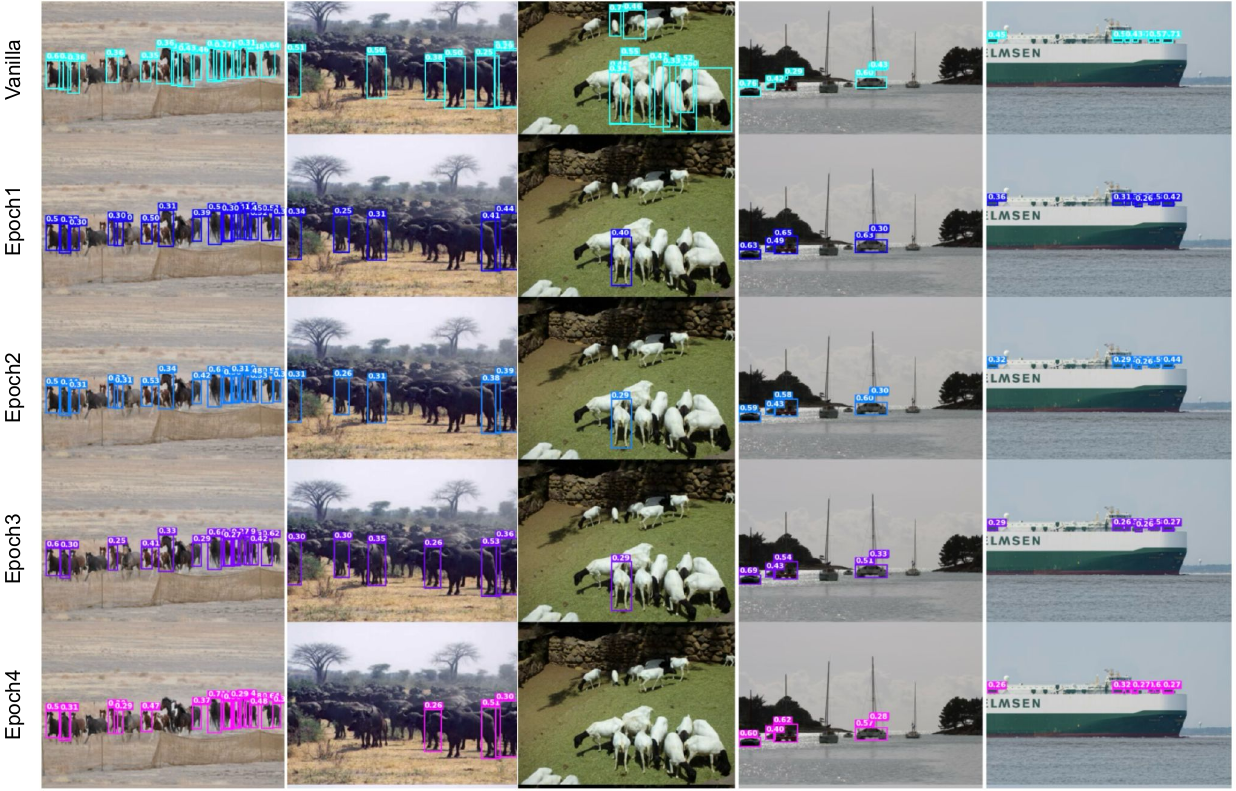} 
    \caption{Qualitative evolution of object detection predictions on representative OoD images selected for their persistent detections across epochs. Rows show predictions from the vanilla model and fine-tuned models across epochs. These examples demonstrate the progressive reduction in hallucination frequency and confidence scores throughout fine-tuning, illustrating how the model becomes increasingly conservative in making predictions on OoD data. Note that most hallucinations in other OoD images are eliminated after the first epoch, making these persistent cases particularly valuable for visualizing confidence evolution trends.}
    \label{fig:ood_qualitative_evolution_epoch}
\end{figure*}

To better illustrate this trend, Fig.~\ref{fig:ood_qualitative_evolution_epoch} presents qualitative examples of OoD scenes with persistent hallucinated detections across multiple epochs.
In the vanilla model, these hallucinations often appear with high confidence.
As fine-tuning progresses, many of these hallucinations disappear entirely; for those that remain, confidence scores are noticeably lower.
Take the middle example involving sheep: the vanilla model detects multiple objects with high confidence (up to 0.78), but by epoch 4, most detections vanish, and the few that remain have scores closer to 0.2–0.3.
This visual progression indicates a consistent refinement of the model’s decision boundaries, making it less likely to produce overconfident predictions on OoD content.


In summary, these confidence-driven trends, both quantitative and qualitative, suggest that fine-tuning with proximal OoD data teaches the model to treat such inputs with increased uncertainty.
This shift away from overconfident predictions plays a central role in reducing hallucinations and improving the model’s overall robustness against OoD inputs.




%% file: sections/7_discussion.tex
\section{DISCUSSION: RESIDUAL HALLUCINATIONS AND OPEN CHALLENGES}\label{sec:discussion}
Despite the substantial reduction in hallucinated detections achieved by our fine-tuning approach and OoD detection integration, certain failure cases persist.
These residual hallucinations highlight deeper limitations of current detection models that cannot be fully resolved by filtering or post-hoc fine-tuning alone.
We categorize these errors into three representative types, each associated with distinct underlying challenges.

\begin{figure*}[t]
    \centering
    \begin{subfigure}{0.32\textwidth}
        \centering
        \includegraphics[width=\textwidth]{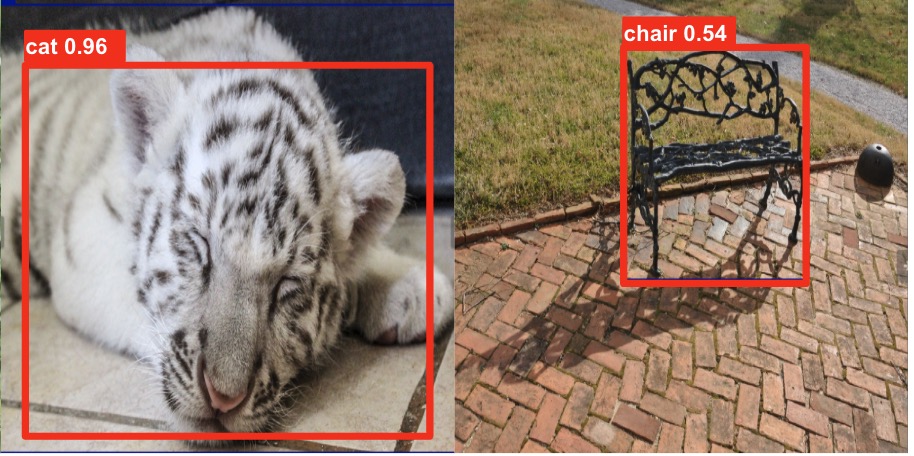}
        \caption{Visual similarity at the class boundary}
        \label{fig:failure_boundary}
    \end{subfigure}
    \hfill
    \begin{subfigure}{0.32\textwidth}
        \centering
        \includegraphics[width=\textwidth]{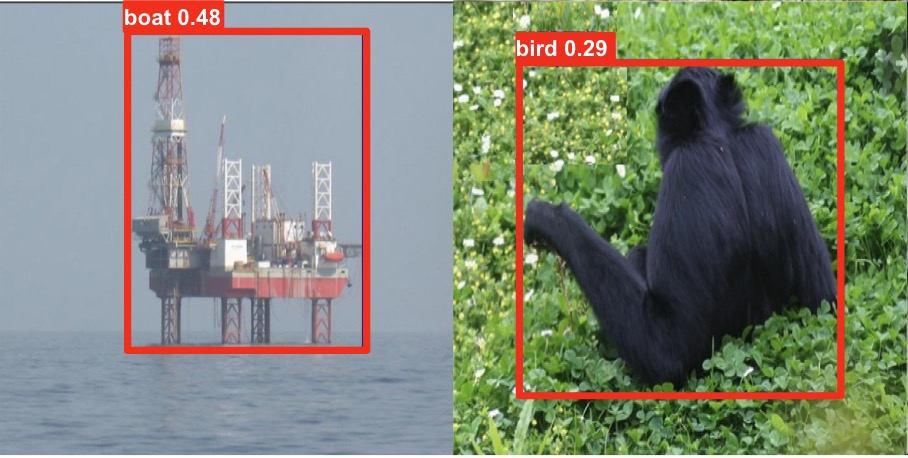}
        \caption{Contextual bias \& spurious correlations}
        \label{fig:failure_contextual}
    \end{subfigure}
    \hfill
    \begin{subfigure}{0.32\textwidth}
        \centering
        \includegraphics[width=\textwidth]{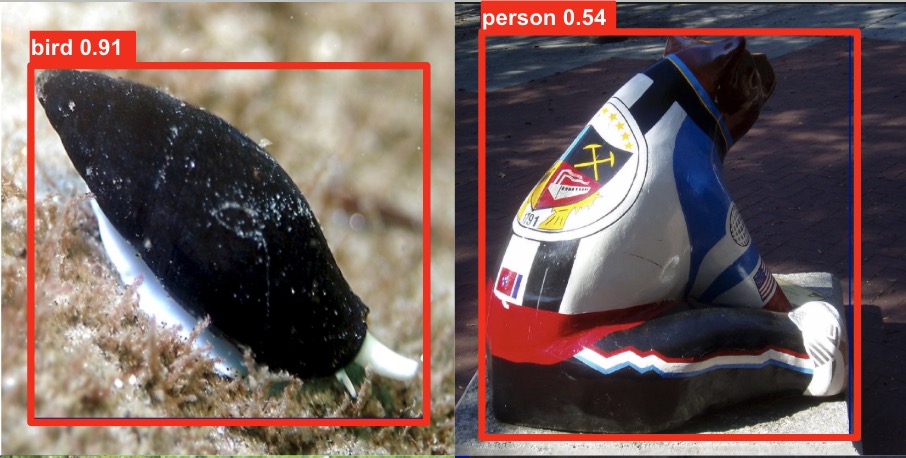}
        \caption{Overgeneralization from geometric cues}
        \label{fig:failure_shape}
    \end{subfigure}
    \caption{Representative failure types in fine-tuned YOLO models with KNN-based OoD detection on the PASCAL VOC benchmark. Each type represents different fundamental challenges: (a) Visual similarity at the class boundary: hallucinations persist due to fine-grained visual similarity, (b) Contextual bias: predictions are dominated by background context rather than object features, and (c) Geometric overgeneralization: coarse shape similarity leads to confusion despite distinctive fine-grained details.}
    \label{fig:failure_cases}
\end{figure*}

\textbf{Visual Similarity at the Class Boundary} Certain hallucinations persist due to fine-grained visual similarity between OoD and in-distribution classes, especially near category boundaries.
For instance, Fig.~\ref{fig:failure_boundary} shows that white tiger cub is misclassified as a cat, and a narrow bench as a chair.
Both cases involve high visual overlap and reflect the absence or coarse granularity of corresponding categories in the training set. 
These errors underscore the difficulty in distinguishing semantically distinct but visually similar objects when class definitions are coarse, a challenge studied in fine-grained recognition \cite{krause2015finegrained} and open-set recognition \cite{bendale2016towards}.
Addressing such ambiguity may require models that better encode part-level visual features and support more flexible or hierarchical category systems.

\textbf{Contextual Bias and Spurious Correlations} In these cases, the model's predictions are heavily influenced by background context rather than the actual object features. 
For instance, as illustrated in Fig.~\ref{fig:failure_contextual}, a monkey sitting on green grass is misclassified as a bird, likely because the model associates the grassy background with birds, neglecting the animal's actual characteristics.
Similarly, an offshore drilling platform is misclassified as a boat, possibly due to its location on the sea and visual resemblance to marine vessels in structure and context.
These errors reflect the model’s over-reliance on background cues and spurious correlations learned during training, leading to hallucinations or misclassifications that ignore foreground semantics.
Such failures are closely tied to longstanding challenges in computer vision, including shortcut learning~\cite{geirhos2020shortcut}, contextual bias~\cite{beery2018recognition}, and the need for causal or disentangled representations that can better separate object identity from scene context.

\textbf{Overgeneralization from Geometric Cues} These errors occur when the model overly relies on coarse geometric features such as silhouette or contour, which leads to misclassification of unfamiliar objects that resemble known categories in shape.
For example, as shown in Fig.~\ref{fig:failure_shape}, a plant with a bird-like posture is mistaken for a bird, and a bear sculpture dressed in clothes is classified as a person.
While these cases could in principle be disambiguated using fine-grained details such as the presence of feathers, facial structure, or texture, the model appears to prioritize shape over finer semantic cues.
Such errors reflect a broader challenge in visual recognition: disentangling superficial shape similarity from true category identity, a problem studied in shape-bias in deep networks~\cite{geirhos2018imagenettrained}, and presents challenges for appearance-based OoD detection, where visually unfamiliar objects may still trigger false positives due to geometric similarity \cite{fort2021exploring, li2020improving}.

\textbf{Summary.} These hallucination errors, spanning class-boundary ambiguity, contextual misguidance, and shape-based overgeneralization, highlight key failure modes in open-world visual recognition.
Addressing these challenges calls for further advances in both computer vision and OoD detection, particularly in developing models that can disentangle object identity from confounding context or shape priors, which is an important direction beyond the scope of this work.

%% file: sections/8_conclusion.tex
\section{CONCLUSION AND FUTURE WORK}\label{sec:conclusion}

This work repositions the task of OoD detection in object detection.
We found that mislabeling in OoD test datasets and anomalies within ID datasets significantly affect evaluation outcomes.
To address this, we developed novel benchmarks to provide more accurate and robust assessment of OoD detection in object detection.
Additionally, we propose a new mitigation paradigm for OoD-induced hallucinations in object detection, which shifts the focus from post hoc filtering to proactive boundary shaping.
This is achieved by fine-tuning on proximal OoD data, allowing the model to treat such inputs as background and learn a more conservative decision boundary.
Future work includes formalizing hierarchical OoD definitions to better generalize proximal examples and define decision boundaries more precisely.
To reduce contextual hallucinations, we plan to explore debiasing strategies that limit over-reliance on background cues.
For shape-based errors, advancing fine-grained representation learning, inspired by shape-bias analysis~\cite{geirhos2018imagenettrained}, is a promising direction.
Additionally, selecting diverse real and synthetic OoD sources remains critical for improving generalization across failure types.
We hope our work’s emphasis on the full OoD mitigation lifecycle inspires new directions for more effective and reliable defense against OoD-induced risks, not only in object detection but also in broader deep learning applications.